\documentclass[12pt]{article}
\RequirePackage{setspace}
    \AtBeginDocument{\setlength\abovedisplayskip{3pt}}
    \AtBeginDocument{\setlength\belowdisplayskip{3pt}}
\usepackage{amsfonts}
\usepackage{bbm}
\usepackage{graphicx}
\usepackage{sectsty}
\sectionfont{\normalsize}
\usepackage{appendix}
\usepackage{float}
\usepackage{url}
\usepackage{caption}
\begin{document}
\title{\Large Concepts and Their Dynamics: A Quantum-Theoretic \\ Modeling of Human Thought}
\author{\normalsize Diederik Aerts \\ 
        \small\itshape \vspace{-0.1 cm}
        Center Leo Apostel for Interdisciplinary Studies and Department  \\
        \small\itshape \vspace{-0.1 cm}
        of Mathematics, Brussels Free University, Brussels, Belgium \\ 
         \small \vspace{-0.1 cm}
        email: \url{diraerts@vub.ac.be} \vspace{0.2 cm} \\ 
             \normalsize   Liane Gabora  \\ 
       \vspace{-0.1 cm} \small\itshape
        Department of Psychology, University of British Columbia \\
       \vspace{-0.1 cm} \small\itshape 
         Kelowna, British Columbia, Canada \\
        \small \vspace{-0.1 cm}
        email: \url{liane.gabora@ubc.ca} \vspace{0.2 cm} \\
         \normalsize Sandro Sozzo  \\ 
       \vspace{-0.1 cm} \small\itshape
        Center Leo Apostel for Interdisciplinary Studies \\
       \vspace{-0.1 cm} \small\itshape
         Brussels Free University, Brussels, Belgium \\
       \vspace{-0.1 cm} \small
        email: \url{ssozzo@vub.ac.be}
        }
\date{}
\maketitle
\vspace{-1cm}
\begin{abstract}
\noindent 
We analyze different aspects of our quantum modeling approach of human concepts, and more specifically focus on the quantum effects of contextuality, interference, entanglement and emergence, illustrating how each of them makes its appearance in specific situations of the dynamics of human concepts and their combinations. We point out the relation of our approach, which is based on an ontology of a concept as an entity in a state changing under influence of a context, with the main traditional concept theories, i.e. prototype theory, exemplar theory and theory theory. We ponder about the question why quantum theory performs so well in its modeling of human concepts, and shed light on this question by analyzing the role of complex amplitudes, showing how they allow to describe interference in the statistics of measurement outcomes, while in the traditional theories statistics of outcomes originates in classical probability weights, without the possibility of interference. The relevance of complex numbers, the appearance of entanglement, and the role of Fock space in explaining contextual emergence, all as unique features of the quantum modeling, are explicitly revealed in this paper by analyzing human concepts and their dynamics.
\end{abstract}

\vspace{-0.1cm}
\noindent
{\bf Keywords:} concept theory, quantum modeling, entanglement, interference, context, emergence, human thought

\vspace{-0.4cm}
\section{Introduction}
\vspace{-0.4cm}
To understand the structure and dynamics of human concepts, how such concepts combine to form sentences, and how meaning is expressed by such combinations, is one of the age-old challenges of scientists studying the human mind. In addition to being a cornerstone for a deeper understanding of human thinking, it is a crucial condition for progress in many fields, including psychology, linguistics, artificial intelligence, and cognitive science. Major scientific issues, such as text analysis, information retrieval and human-computer interaction, are likewise reliant on a deeper insight into human concepts. To achieve such insight has proved to be deceptively difficult, however. It was shown, for example, that concept combinations cannot be modeled properly by using fuzzy set theory (Osherson \& Smith, 1981), and that rules of classical logic are violated when membership weights of combined concepts are related to membership weights of component concepts (Hampton, 1988a,b). Much effort has been devoted to these matters, but very few substantial results have been obtained. Actually, none of the traditional concept theories provides a satisfactory model for the dynamics of concepts and the way they combine.

More than a decade ago, we started to recognize the presence of typical quantum stuctures and effects in the dynamics of concepts and their combinations. More specifically, we identified quantum entanglement, showed the presence of contextuality and interference of a genuine quantum type, and proved the appearance of emergence related to quantum superposition (Aerts, 2009a; Aerts et al., 2000; Aerts \& Gabora, 2005a,b; Aerts \& Sozzo, 2011; Gabora \& Aerts 2002). Unlike traditional concept models, the models of human concepts and their dynamics that we developed by making explicit use of the mathematical formalisms of quantum theory perform well at modeling data gathered through different concept experiments, notably in concept combination studies. What has been put forward to a lesser extent is `why' this quantum approach works. The goal of this paper is to analyze and attempt to explain what is going on `underground', so to speak, that makes the approach so effective. In addition to this explanatory focus of the present article, we will present a review of the most important elements of our quantum modeling approach. 
We also refer to the introductory paper Wang et al. (2012) of the special issue where the present article appears, which provides a general overview, pointing out the potential of quantum modeling for cognitive processes.
 
\vspace{-0.4cm}
\section{Axiomatics, States, Contexts, Gradedness and Fuzziness  \label{glimmers}}
\vspace{-0.4cm}
It can be seen as a very bold move to start using the quantum formalism to model concepts and their dynamics, particularly considering the general lack of understanding of quantum physics itself, expressed colorfully by one of the greatest quantum physicists, Richard Feynman, when he asserted that, `Nobody understands quantum mechanics' (Feynman, 1967). On the other hand, due to this very hard struggle of physicists to `understand' the quantum world, many aspects of quantum theory have been investigated in great depth over the years, laying the foundations for a strict mathematical and axiomatic framework that also offers a very profound operational basis (Jauch, 1968; Mackey, 1963; Piron, 1976). For several decades, the activities of our Brussels research group followed the axiomatic and operational approaches to quantum physics (Aerts, 1982a, 1986, 1999; Aerts \& Durt, 1994, Aerts, Coecke \& Smets, 1999; Aerts et al., 1997), identifying how specific macroscopic situations -- i.e. not necessarily situations of quantum particles in the micro-world -- can be modeled more adequately by a quantum-like formalism than by a classical formalism (Aerts, 1982b, 1991; Aerts \& Van Bogaert, 1992; Aerts et al., 1993, 2000). These ventures of using quantum structures to model macroscopic situations have played an essential role in the growing evidence that a quantum description would be appropriate for situations in human cognition. Indeed, we then applied these insights to model a typical situation of human decision-making (Aerts \& Aerts, 1995).

With respect to applying the mathematical formalism of quantum mechanics to concepts and their dynamics, one of the major conceptual steps we took was to consider a concept fundamentally as an `entity in a specific state', and not, as in the traditional approaches, as a `container of instantiations'. This container view of a concept remains present in the traditional approaches, even if fuzzy structures are introduced. After all, a fuzzy set is a set, i.e. a container, even if it is a fuzzy one. We started to conceive of a concept in this manner in the years 1998-1999, and first explored the notion of entanglement -- admittedly still on a predominantly formal level --, by proposing a situation of change of state, a collapse, from a more abstract form of a concept to a more concrete one, violating Bell's inequalities (Aerts et al., 2000, Gabora 2001). Recent experiments confirmed such a violation of Bell's inequalities (Aerts \& Sozzo, 2011), and we will explore this in detail in section \ref{entanglement}. The notion of `state of a concept' and the corresponding notion of `change of state or collapse' proved to be very valuable when we investigated connections between our quantum-inspired approach and traditional concept theories.

In the oldest view on concepts, going back to Aristotle, and now referred to as `the classical view', all instances of a concept share a common set of necessary and sufficient defining properties. Wittgenstein (1953) already noted that there is a compelling incongruity between this classical view and how people actually use concepts. The critical blow to the classical view, however, was given by Rosch's work on color (1973). It was shown that colors do not have any particular criterial attributes or definite boundaries, and instances differ with respect to how typical or exemplary they are of a category. This led to the formulation of the `prototype theory', according to which concepts are organized based on family resemblances and consist of characteristic rather than defining features, which are weighted in the definition of the prototype (Rosch, 1978, 1983). Rosch showed that subjects rate concept membership as graded, with the degree of membership of an instance corresponding to the conceptual distance from the prototype. Another strength of the prototype theory is that it can be mathematically formulated and empirically tested. By calculating the similarity between the prototype and a possible instance of a concept, across all salient features, one arrives at a measure of conceptual distance between the instance and the prototype. It is in this respect that Osherson and Smith (1981) identified a fundamental problem concerning prototype theory and the combinations of concepts, which has come to be known as the `Pet-Fish Problem'. If the concept of {\it Pet-Fish} is the conjunction of the concepts {\it Pet} and {\it Fish}, it should follow from fuzzy set theory -- where standard connectives for conjunction involve typicality values that are less than or equal to each of the typicality values of the components -- that the typicality of {\it Guppy} is not higher for {\it Pet-Fish} than for either {\it Pet} or {\it Fish}. In reality, while people rate {\it Guppy} neither as a typical {\it Pet} nor as a typical {\it Fish}, they do rate it as a highly typical {\it Pet-Fish}.  This phenomenon of the typicality of a conjunctive concept being greater than that of either of its constituent concepts has been defined as the `Guppy Effect'. It defies the standard fuzzy set modeling of the behavior of typicality with respect to conjunction. A similar effect has been proven to exist for membership weights of exemplars. Hampton (1988a,b) showed that people estimated membership in such a way that the membership weight of an exemplar of a conjunction of concepts, calculated as the relative frequency of membership estimates, is higher than the membership weights of this exemplar for one or both of the constituent concepts. This phenomenon is referred to as `overextension'. `Double overextension' is also an experimentally established phenomenon. In this case, the membership weight of the exemplar for the conjunction of concepts is higher than the membership weights for `both' constituent concepts (Hampton, 1988a,b, 1997).
The problems encountered with composition, the graded nature of exemplars, the problematic fuzzy nature of typicality and the probabilistic nature of membership weight, are factors that all have been accounted for in our quantum approach, as we will systematically show in the different sections of the present article.

In our quantum approach, a context is modeled mathematically as a measurement, and hence it changes the state of a concept  in the way a measurement in quantum theory changes the state of a quantum entity (Aerts \& Gabora, 2005a; Gabora \& Aerts, 2002). For example, in our modeling of the concept {\it Pet}, we considered the context $e$ expressed by {\it Did you see the type of pet he has? This explains that he is a weird person}, and found that when participants in an experiment were asked to rate different exemplars of {\it Pet}, the scores for {\it Snake} and {\it Spider} were very high in this context. In our approach, this is explained by introducing different states for the concept {\it Pet}. We call `the state of {\it Pet} when no specific context is present', its ground state $\hat p$. The context $e$ then changes the ground state $\hat p$ into a new state $p_{weird\ person\ pet}$. Typicality, in our approach, is an observable semantic quantity, which means that it takes different values in different states. Hence, in our approach the typicality variations as encountered in the guppy effect are due to changes of state of the concept {\it Pet} under influence of a context. More specifically, the conjunction {\it Pet-Fish} is {\it Pet} under the context {\it Fish}, in which case the ground state $p$ of {\it Pet} is changed into a new state $p_{Fish}$. The typicality of {\it Guppy}, being an observable semantic quantity, will be different depending on the state, and this explains the high typicality of {\it Guppy} in the state $p_{Fish}$ of {\it Pet}, and its normal typicality in the ground state $p$ of {\it Pet} (Gabora \& Aerts 2002).

We developed this approach in a formal way, and called the underlying mathematical stucture a State Context Property System, abbreviated SCoP (Aerts \& Gabora 2005a). To build SCoP, for an arbitrary concept $S$, we introduce three sets, namely the set $\Sigma$ of states, denoting states by $p, q, \ldots$, the set ${\mathcal M}$ of contexts, denoting contexts by $e, f, \ldots$, and the set ${\mathcal L}$ of properties, denoting properties by $a, b, \ldots$. The `ground state' $\hat{p}$ of the concept $S$ is the state where $S$ is not under the influence of any particular context. Whenever $S$ is under the influence of a specific context $e$, a change of the state of $S$ occurs. In case $S$ was in its ground state $\hat{p}$, the ground state changes to a state $p$. The difference between states $\hat{p}$ and $p$ is manifested, for example, by the typicality values of different exemplars of the concept, and the applicability values of different properties being different in the two states $\hat{p}$ and $p$. Hence, to complete the mathematical construction of SCoP, also two functions $\mu$ and $\nu$ are introduced. The function $\mu: \Sigma \times {\mathcal M} \times \Sigma \longrightarrow [0, 1]$ is defined such that $\mu(q,e,p)$ is the probability that state $p$ of concept $S$ under the influence of context $e$ changes to state $q$ of concept $S$. The function $\nu: \Sigma \times {\mathcal L} \longrightarrow [0, 1]$ is defined such that $\nu(p,a)$ is the weight, or normalization of applicability, of property $a$ in state $p$ of concept $S$. With these mathematical structures and tools the SCoP formalism copes with both `contextual typicality' and `contextual applicability'.

We likewise built an explicit quantum representation in a complex Hilbert space of the data of the experiment on {\it Pet} and {\it Fish} and different states of {\it Pet} and {\it Fish} in different contexts explored in Aerts \& Gabora (2005a), as well as of the concept {\it Pet-Fish} (Aerts \& Gabora 2005b). In recent work, we have complemented this representation by considering the quantum effect of interference which also takes place in the situation of the guppy effect (Aerts et al. 2012).

Our approach can be interpreted in a rather straightforward way as a generalization of prototype theory, which explicitly integrates context, unlike standard prototype theory. What we call the ground state of a concept can thus be seen as the prototype of this concept. However, in our approach, any context will change this ground state into a new state. Consequently, when the concept is in this new state, the standard prototype ceases to play the role of prototype. An intuitive way of understanding our approach as a generalization of prototype theory, with maximum use of the wording of prototype theory, would be to consider this new state as a new `contextualized prototype'. Hence, concretely, {\it Pet}, when combined with {\it Fish}, has a new contextualized prototype, something which could be called {\it Pet} `contextualized by Fish'. Of course, the word `prototype' may no longer adequately express this new state, but thinking of the new state as a `contextualized prototype' aids in grasping the type of generalization our approach constitutes with respect to prototype theory. Hence, the positioning of our quantum-inspired approach to prototype theory is rather clear, because we put forward a prototype-like theory that is capable of fully describing the presence and influence of context. 

By specifying the relation of our approach to the two other main traditional concept theories, `exemplar theory' and `theory theory', we can clarify why our approach is more than the contextual generalization of prototype theory. In exemplar theory, a concept is represented by, not a set of defining or characteristic features, but a set of salient instances of it stored in memory (Nosofsky, 1988, 1992). In theory theory concepts take the form of `mini-theories' (Murphy \& Medin, 1985) or schemata (Rumelhart \& Norman, 1988) identifying the causal relationships amongst properties. These two approaches reveal their focus on one specific aspect of prototype theory, namely the situation where a concept is considered to be determined by its set of characteristic rather than defining features. This means that both theories have mainly been preoccupied with the question, `What is it that predominantly determines a concept?' Although the importance of this question is self-evident, it has not been the main issue we have focused on in our approach. Transposed to our approach, this question would read as follows, `What is it that predominantly determines the state of a concept'. The main preoccupation of our approach has been to propose a theory (i) with a well-defined ontology, i.e. we consider a concept to be an entity capable of different modes of being with respect to how it influences measurable semantic quantities such as typicality or membership weight, and we call these modes states; and (ii) that manages to produce theoretical models capable of fitting experimental data on these measurable semantic quantities. We seek to achieve (i) and (ii) mainly independently of the question that is the focus of `exemplar theory' and `theory theory', although we do acknowledge that this is an important aspect. More concretely, and in accordance with the results of investigations into the question of `What predominantly determines a concept' as far as prototype theory, exemplar theory and theory theory are concerned, we do believe that all approaches are valid in part. The state of a concept, i.e. its capability of influencing the values of measurable semantic quantities such as typicality and membership weight, is influenced by the set of its characteristic features, but also by salient exemplars stored in memory, and in a considerable number of cases -- where more causal aspects are at play -- mini-theories might be appropriate to express this state. For our approach, it is important that `the state exists and is what gives rise to the values of the measurable semantic quantities', albeit probabilistically, which in the presence of quantum structure means that potentiality and uncertainty appear even if the state is completely known.

At this point, we want to underline specifically this important difference between the quantum probability in our quantum approach and general classical probability or fuzziness as used in the traditional approaches. The modeling of uncertainty and potentiality in quantum probability theory is profoundly different from that in classical probability theory, i.e. the theory developed from reflections about chance and games (Laplace, 1820) and formulated later axiomatically (Kolmogorov, 1933). Note that it is not the interpretation we refer to but the mathematical structure of quantum probability itself, which is different from that of classical probability. The guppy effect might still be regarded as not conclusively pointing towards the presence of quantum structure, in the sense that many types of connective can be put forward in fuzzy set theory, including one that copes with the strange behavior encountered in the guppy effect. The situation is more serious for the deviations of membership weights identified by Hampton (1988a). Relying on investigations in quantum probability, it can be proven that this cannot be modeled within a classical probability theory. More concretely, the membership weights $\mu(A), \mu(B)$ and $\mu(A\ {\rm and}\ B)$ can be represented within a classical probability model if and only if the following two requirements are satisfied (Aerts, 2009a, theorem 3)
\begin{eqnarray} \label{mindeviation}
\mu(A\ {\rm and}\ B)-\min(\mu(A),\mu(B))=\Delta_c\le 0 \\ \label{kolmogorovianfactorconjunction}
0 \le k_c=1-\mu(A)-\mu(B)+\mu(A\ {\rm and}\ B)
\end{eqnarray}
where we called $\Delta_c$ the `conjunction rule minimum deviation', and $k_c$, the `Kolmogorovian conjunction factor'. Equations (\ref{mindeviation}) and (\ref{kolmogorovianfactorconjunction}) can be intuitively grasped by noticing that the former is a consequence of monotonicity of probabilities, while the latter derives from their additivity. Hampton (1988a) considered, for example, the concepts {\it Food} and {\it Plant} and their conjunction {\it Food and Plant}. He then conducted tests to measure the extent to which participants decided whether a certain exemplar was or was not a member of each concept. In the case of the exemplar {\it Mint}, the outcome -- the relative frequency of membership -- was 0.87 for the concept {\it Food}, 0.81 for the concept {\it Plant}, and 0.9 for the conjunction {\it Food and Plant}. This means that participants found {\it Mint} to be more strongly a member of the conjunction {\it Food and Plant} than they found it to be a member of either of the two component concepts {\it Food} and {\it Plant}. Also, with $A$ being {\it Food} and $B$ being {\it Plant}, inequality (\ref{mindeviation}) is violated, because $\Delta_c=0.09\not\le0$. Hence from theorem 3 (Aerts, 2009a) follows that there is no classical probability representation for these data. The issue of hidden variables in the foundations of quantum physics is related to these investigations into the nature of quantum probability theory. In this sense, a very general geometric method was devised to see whether experimental data could be modeled within a classical probability theory. In Aerts et al. (2009), we confronted all Hampton (1988a) data with this geometrical method, and found an abundance of data that could not be modeled in a classical probability theory.

The typical differences between quantum probability theory and classical probability theory immediately suggest that the way {\it Mint} violates inequality (\ref{mindeviation}) involves a genuine quantum effect, not reproducible by even the most complex classically based theoretical framework.

\vspace{-0.4cm}
\section{The Quantum Realm \label{inklings}}
\vspace{-0.4cm}
We will now investigate parts of what we called `the underground' in the introduction, i.e. the place where quantum theory is at work, and produces effects fundamentally different from effects due to classical probability and fuzziness. Compared to quantum theory, classical probabilistic theories, and also fuzzy set theories, may be said to be merely scratching the surface of things. Here, `things' refer to a `set of situations involving entities, the performance of experiments and the collection of outcomes'. Classical probability theory -- and this is somewhat less clear for fuzzy set theory, but nonetheless true -- has developed a very general approach to the modeling of such a `set of situations involving entities, the performance of experiments and the collection of outcomes'. Its narrow and limited quality is apparent, however, from the fact that in the end it is a positive number between zero and one, the probability, that plays the primary role in what is aimed to be expressed regarding this `set of situations involving entities, the performance of experiments and the collection of outcomes'. A mathematical calculus was developed showing the probability as the limit of a fraction, viz. the number of desired outcomes of the experiment divided by the number of experiments done. What could be a more general way to connect mathematics with reality? 

Quantum theory as a way to describe a `set of situations involving entities, the performance of experiments and the collection of outcomes' arose in physics, and the phenomenon of `interference' has played an important role in its genesis. Interference is the phenomenon of interacting waves, and both positive and negative numbers are key to their interference pattern. A wave is best modeled by regarding its crests as positive lengths away from the average and its troughs as negative lengths away from the average. When two waves interact, the crests of the one, when colliding with the troughs of the other, will partly be annihilated, giving rise to a specific pattern. This is comparable to how part of the value of a positive number is annihilated when the number is summed with a negative number. The pattern following from this partial annihilation between crests and troughs is the interference pattern.

Interference is a good example of a phenomenon that cannot be described by only focusing on positive numbers. It may of course be possible that `probabilities' and `fuzziness' in essence do not interfere. We would be tempted to believe this if we considered, for example, the relative frequency definition of probability. It is an experimental fact that in the micro-world probabilities interfere. This is mathematically taken into account in quantum theory through the use of complex numbers rather than positive real numbers to express the probabilities. More specifically, it is the square of the absolute value of this complex number that gives rise to the probability.
There is also interference when probabilities are considered in relation to human decisions.
We believe that, in depth, this is due to both realms -- the one of quantum particles in the micro-world and the one of conceptual structures in semantic space -- being realms of genuine `potentialities', not of the type of a `lack of knowledge of actualities'. This insight also led us to present a quantum model for a human-decision process (Aerts \& Aerts, 1995), since as a rule human decisions are made in a state of genuine potentiality, which is not of the type of a lack of knowledge of an actuality. The following example serves to illustrate this. In Aerts \& Aerts (1995), we considered a survey including the question `are you a smoker or not'. Suppose that of the total of 100 participants 21 said `yes' to this question. We can then attribute $0.21$ as the probability of finding a smoker in this sample of participants. However, this probability is obviously of the type of a `lack of knowledge about an actuality', because each participant `is' a smoker or `is not' a smoker. Suppose that we now consider the question `are you for or against the use of nuclear energy?' and that 31 participants say they are in favor. In this case, the resulting probability, i.e. $0.31$, is `not' of the type of `lack of knowledge about an actuality'. Indeed, since some of the participants had no opinion about this question before the survey, the outcome was influenced by the `context' at the time the question was asked, including the specific conceptual structure of how the question was formulated. This is how `context' plays an essential role whenever the human mind is concerned with the outcomes of experiments. It can be shown that the first type of probability, i.e. the type that models a `lack of knowledge about an actuality', is classical, and that the second type is non-classical (Aerts, 1986).

Let us now illustrate how we model interference for Hampton's (1988a) membership test by using the quantum formalism. We consider two concepts $A$ and $B$, and their conjunction $A\ {\rm and}\ B$, and an exemplar. Hampton (1988a) measured membership weights, i.e. probabilities $\mu(A)$, $\mu(B)$ and $\mu(A\ {\rm and}\ B)$, for this exemplar to be a member of $A$, $B$ and $A\ {\rm and}\ B$.
The equation that describes a general quantum modeling -- we refer to Aerts (2009a), section 1.7, for its derivation, and mention that conjunction has been treated differently in Franco (2009)  --  is the following
\begin{equation} \label{membershipweightinterference}
\mu(A\ {\rm and}\ B)=m^2\mu(A)\mu(B)+n^2({\mu(A)+\mu(B) \over 2}+\Re\langle A|M|B\rangle)
\end{equation}
where the numbers $0 \le m^2 \le 1$ and $0 \le n^2 \le 1$ are convex coefficients, hence we have $m^2+n^2=1$, and $\Re\langle A|M|B\rangle$ is the interference term.

In this interference term, the mathematical objects appear that are fundamental to the quantum formalism, namely unit vectors of a complex Hilbert space ($\langle A|$ and $|B\rangle$ are such unit vectors), and orthogonal projection operators on this complex Hilbert space ($M$ is such an orthogonal projection operator). The expression $\langle A|M|B\rangle$ is the inner product in this complex Hilbert space, and this is a complex number. $\Re\langle A|M|B\rangle$ is the `real part' of this complex number. For readers not acquainted with the quantum formalism, we have added in appendix A an explanation of the mathematics needed for our use of quantum theory.

Let us first remark that $\mu(A)\mu(B)$ is what would be expected for the conjunction in the case of a classical probability. Hence, when $m=1$ and $n=0$, we are in this `classical probability' situation. Since $\mu(A)\mu(B) \le \mu(A)$ and $\mu(A)\mu(B) \le \mu(B)$, we have $\mu(A)\mu(B) \le {\mu(A) + \mu(B) \over 2}$. Suppose that we do not have interference, and hence $\Re\langle A|M|B\rangle=0$, then (\ref{membershipweightinterference}) expresses that $\mu(A\ {\rm and}\ B)\in [\mu(A)\mu(B), {\mu(A)+\mu(B) \over 2}]$. Hence for values of $\mu(A\ {\rm and}\ B)$ outside this interval a genuine interference contribution is needed for (\ref{membershipweightinterference}) to have a solution. We introduce
\begin{equation}
0\le f_c=\min({\mu(A)+\mu(B) \over 2} - \mu(A\ {\rm and}\ B),\mu(A\ {\rm and}\ B)-\mu(A)\mu(B))
\end{equation}
as the criterion for a possible solution without the need for genuine interference. Let us consider again the example of {\it Mint}, with respect to {\it Food}, {\it Plant} and {\it Food and Plant} and its data taken from Hampton (1988a). We have $f_c=-0.06$, and hence for {\it Mint} the quantum interference term needs to attribute for a solution of equation (\ref{membershipweightinterference}) to be possible.
We showed in Aerts (2009a), section 1.5, that a solution can be found in a three-dimensional complex Hilbert space, such that 
\begin{equation} \label{interferencecosinus}
\Re\langle A|M|B\rangle=\sqrt{(1-\mu(A))(1-\mu(B))}\cos\beta
\end{equation}
where $\beta$ is the interference angle, and the unit vectors $|A\rangle$ and $|B\rangle$ representing the states of the concepts $A$ and $B$ in the canonical basis of this three-dimensional complex Hilbert space are as follows
\begin{eqnarray} \label{vectorA}
\!\!\!\! |A\rangle&\!\!=\!\!&(\sqrt{\mu(A)},0,\sqrt{1-\mu(A)}) \\ \label{vectorB}
\!\!\!\! |B\rangle&\!\!=\!\!&e^{i\beta}(\sqrt{(1-\mu(A))(1-\mu(B)) \over \mu(A)},\sqrt{\mu(A)+\mu(B)-1 \over \mu(A)},-\sqrt{1-\mu(B)}) \\ \label{anglebeta}
\!\!\!\! \beta&\!\!=\!\!&\arccos({{2 \over n^2}(\mu(A\ {\rm and}\ B)-m^2\mu(A)\mu(B))-\mu(A)-\mu(B) \over 2\sqrt{(1-\mu(A))(1-\mu(B))}}) 
\end{eqnarray}
To construct a solution for {\it Mint}, we take $m^2=0.3$ and $n^2=0.7$, and find the value for $\beta$, using (\ref{anglebeta}), $\beta=50.21^\circ$. We can see in the vector $|B\rangle$ representing concept $B$ that a complex number plays an essential role, namely $e^{i\beta}$, which is present in all components of the vector. This is the root of the interference, and hence the deep reason that the membership weight of {\it Mint} for {\it Food and Plant} can be bigger than the membership of {\it Mint} for {\it Food} and the membership of {\it Mint} for {\it Plant}.

To show this explicitly in the above Hilbert space model would be technically sophisticated -- see Aerts (2009a) for an explicit discussion. For this reason, we will use just two complex numbers to illustrate directly `how they allow interference to take place'. Consider two complex numbers $ae^{i\alpha}$ and $be^{i\beta}$, both of absolute values, $a$ and $b$, respectively, smaller than or equal to 1, such that the squares of absolute values, hence $a^2$ and $b^2$, can represent probabilities. Suppose they represent probabilities $\mu(A)$ and $\mu(B)$ of disjoint events $A$ and $B$, such that $A \cap B=\emptyset$. In classical probability, the probability of the joint event $\mu(A \cup B)$ is in this case equal to the sum $\mu(A)+\mu(B)$. In quantum theory, the complex numbers need to be summed and subsequently the probability is calculated from this sum of complex numbers by squaring its absolute value. Hence, we get $\mu(A \cup B)=(ae^{i\alpha}+be^{i\beta})^*(ae^{i\alpha}+be^{i\beta})=(ae^{-i\alpha}+be^{-i\beta})(ae^{i\alpha}+be^{i\beta})=a^2+b^2+abe^{i(\beta-\alpha)}+abe^{-i(\beta-\alpha)}=a^2+b^2+2ab\cos(\beta-\alpha)$, where we have used Euler's equation $\cos\phi+i\sin\phi=e^{i\phi}$, for an arbitrary angle $\phi$. This shows that
\begin{equation}
\label{barecomplexnumbers}
\mu(A \cup B)=\mu(A)+\mu(B)+2\sqrt{\mu(A)\mu(B)}\cos(\beta-\alpha)
\end{equation}
Consider the similarity between equations (\ref{barecomplexnumbers}) and (\ref{membershipweightinterference}), certainly following substitution (\ref{interferencecosinus}) in (\ref{membershipweightinterference}). By this we intend to show that interferences, and hence the deviations Hampton (1988a) measured using membership weights, are directly due to the use of complex numbers.

Having put forward the hypothesis that interference on the level of probabilities is connected with the presence of potentiality, we cannot resist the temptation to finish the section by pointing out that the invention of complex numbers was guided by potentiality as well. Back in 1545, when even negative numbers were not completely allowed as genuine numbers, the Italian mathematician Gerolamo Cardano introduced `imaginary numbers' in an attempt to find solutions to cubic equations (Cardano, 1545). What is amusing is that Cardano used these numbers in intermediate steps, which is why they only played the role of `potential numbers'. He eliminated them once he had found real number solutions. While he called them `sophistic' and `a mental torture', he was also fascinated by them. Independently of their practical use in his work on cubic equations, he put forward, for example, the problem of finding two numbers that have a sum of 10 and a product of 40. A solution is of course the complex numbers $5\pm i\sqrt{15}$. Cardano wrote down the solution, and commented that `this result is as subtle as it is useless'. Cardano would certainly be amazed and delighted to know that quantum interference happens exactly the way in which 10 can be the sum of two numbers whose product is 40. We can see this even explicitly if we use the goniometric representation. Then the two numbers are $\sqrt{40}e^{i\alpha}$ and $\sqrt{40}e^{-i\alpha}$ and their sum  $2\sqrt{40}\cos\alpha$. For an interference angle $\alpha=37.76^\circ$, we find the sum equal to 10.

Is quantum theory the final theory for what we called `the underground'? That is not certain. The aspects that can be additionally modeled by quantum theory, and cannot be modeled by classical probability or fuzziness, such as `interference', `entanglement', and `emergence', continue to be the subject of constant investigation with a view to gaining a better understanding. Hence, it is very well possible that new and even more powerful `underground theories' will arise, substituting quantum theory, and generalizing it. Quantum axiomatics is a domain of research actively engaged in investigating operational foundations and generalizations of quantum theory.

In the next section, we turn to one of the very intriguing aspects of quantum theory, namely entanglement. We will show experimentally that the way entanglement takes place spontaneously when two concepts combine is very similar to how it takes place in the quantum world.

\vspace{-0.4cm}
\section{Entanglement\label{entanglement}}
\vspace{-0.4cm}
Entanglement is tested by Bell's inequalities, which were historically introduced in physics with respect to the problem of completeness of quantum mechanics. Indeed, quantum theory provides predictions that are correct and confirmed by experiments, but these predictions are generally only probabilistic, i.e. they refer to the statistics of outcomes of repeated measurements. 
Hence the question arose whether it would be possible to introduce
`hidden variables' which, once specified together with the state of the entity, would allow one to predict the outcomes of any measurement performed on the entity itself. In this perspective, quantum probabilities would just formalize the lack of knowledge about these underlying `hidden variables'. There is in physics a well-known class of inequalities, whose prototype is the Bell inequality (Bell, 1964), which hold under reasonable physical assumptions (`local realism') and yet are violated by quantum mechanics in specific physical situations in which different inequalities (`quantum inequalities') hold. 
We briefly summarize the reasoning that led to the deduction of one of these inequalities in the framework of a hidden variable research program (Clauser et al., 1969).   

The term `local realism' has been traditionally used to denote the join of the assumptions of `realism', i.e. the values of all observables of a physical system in a given state are predetermined for any measurement context, and `locality', i.e. if measurements are made at places remote from one another on parts of a physical system which no longer interact, the specific features of one of the measurements do not influence the results obtained with the others.

The standard procedures leading to the Clauser-Horne-Shimony-Holt variant of Bell's inequality can then be resumed as follows. Let $U$ be a composite entity made up of the far-away subentities $S$ and $T$, and let $A(a)$ and $B(b)$ be two dichotomic observables of $S$ and $T$, depending on the experimentally adjustable parameters $a$ and $b$, respectively, and taking either value $-1$ or $+1$. 

By assuming `realism', the expectation value of the product of the dichotomic observables $A(a)$ and $B(b)$ in a state $p$ is
\begin{equation}
E(A,B)=\int_{\Lambda} A(\lambda,a) B(\lambda,b) \rho(\lambda) d\lambda
\end{equation}
where $\lambda$ is a `deterministic hidden variable' whose value ranges over the measurable space $\Lambda$ when different examples of the entity $U$ are considered, while $\rho$ is a probability density on $\Lambda$ (determined by $p$, $A(a)$ and $B(b)$). Finally, $A(\lambda, a), B(\lambda, b)=\pm 1$ are the values of the dichotomic observables $A(a)$ and $B(b)$. If one also introduces the dichotomic observables $A(a')$ and $B(b')$ of $S$ and $T$, respectively, one gets similar formulas for the expectation values $E(A,B')$, $E(A',B)$ and $E(A',B')$. Therefore, one obtains, by assuming `locality', the following Clauser-Horne-Shimony-Holt inequality (Clauser et al., 1969)
\begin{equation}
-2 \le E(A',B')+E(A',B)+E(A,B')-E(A,B)\le 2
\end{equation}  
It is then well known that there are examples in which the Clauser-Horne-Shimony-Holt inequality is violated by the expectation values predicted by quantum mechanics for suitable choices of the parameters $a,b,a',b'$. According to the standard interpretation of quantum mechanics this implies that `realism' and `locality' cannot both hold in quantum mechanics. As long as the inequality was not tested experimentally, views of the issue differed. Some physicists, for example, surmised that a failure of the theoretical framework of quantum theory might be detected for the first time. However, experimental tests confirmed the predictions of quantum mechanics to violate Bell's inequality (Aspect et al., 1982). This definitely changed the attitude of many physicists, in that they now started to take more seriously also the deeply strange aspects of `quantum reality'. The situation of quantum theory that gives rise to the violation of Bell's inequality is called `entanglement' (see appendix A). It corresponds to a specific state, called an `entangled state', in which two entities can be prepared. The presence of such entangled states for two entities indicates that it is no longer possible to consider the two entities as existing independently, because they have essentially merged into one entity that behaves as an undivided whole.

We will show in this section that two simple concepts that are combined give rise to the violation of Bell's inequalities, which proves that concepts combine in such a way that they are entangled. In this section we put forward the results of the experiment we performed, showing how it violates Bell's inequalities, and we will analyze its meaning in section \ref{emergence}.

We consider the concept {\it Animal} and the concept {\it Acts}, and join them to form the conceptual combination {\it The Animal Acts}. To bring about a situation that is necessary to formulate Bell's inequalities, we have to consider experiments on {\it Animal} combined with experiments on {\it Acts}, hence joint experiments on {\it The Animal Acts}, and measure the expectation values of possible outcomes. Bell's inequalities are indeed formulated by means of the expectation values of such joint experiments. 

Let us introduce the experiments $A$ and $A'$ on {\it Animal}, and $B$ and $B'$ on {\it Acts}. Experiment $A$ consists in participants being asked to choose between two exemplars, {\it Horse} and {\it Bear}, as possible answers to the statement `is a good example of {\it Animal}'. Experiment $A'$ consists in asking the participants to choose between the two exemplars {\it Tiger} and {\it Cat}, as possible answers to the statement `is a good example of {\it Animal}'. Experiment $B$ asks the participants to choose between the exemplars {\it Growls} and {\it Whinnies}, as possible answers to the statement `is a good example of {\it Acts}'. Experiment $B'$ considers the choice between the exemplars {\it Snorts} and {\it Meows} regarding the statement `is a good example of {\it Acts}'.

Next we consider the coincidence experiments $AB$, $A'B$, $AB'$ and $A'B'$ for the combination {\it The Animal Acts}. More concretely, in the experiment $AB$, to answer the question `is a good example of {\it The Animal Acts}', participants will choose between the four possibilities (1) {\it The Horse Growls}, (2) {\it The Bear Whinnies} -- and if one of these is chosen we put $E(AB)=+1$ -- and (3) {\it The Horse Whinnies}, (4) {\it The Bear Growls} -- and if one of these is chosen we put $E(AB)=-1$. In the coincidence experiment, $A'B$ participants, to answer the question `is a good example of {\it The Animal Acts}', will choose between (1) {\it The Tiger Growls}, (2) {\it The Cat Whinnies} -- and in case one of these is chosen we put $E(A'B)=+1$ -- and (3) {\it The Tiger Whinnies}, (4) {\it The Cat Growls} -- and in case one of these is chosen we put $E(A'B)=-1$. In the coincidence experiment, $AB'$ participants, to answer the question `is a good example of {\it The Animal Acts}', will choose between (1) {\it The Horse Snorts}, (2) {\it The Bear Meows} -- and in case one of these is chosen, we put $E(AB')=+1$ -- and (3) {\it The Horse Meows}, (4) {\it The Bear Snorts} -- and in case one of these is chosen we put $E(AB')=-1$. Finally, in the coincidence experiment, $A'B'$ participants, to answer the question `is a good example of {\it The Animal Acts}', will choose between (1) {\it The Tiger Snorts}, (2) {\it The Cat Meows} -- and in case one of these is chosen we put $E(A'B')=+1$ -- and (3) {\it The Tiger Meows}, (4) {\it The Cat Snorts} -- and in case one of these is chosen we put $E(A'B')=-1$.

We can now evaluate the expectation values $E(A', B')$, $E(A', B)$, $E(A, B')$ and $E(A,B)$ associated with the coincidence experiments $A'B'$, $A'B$, $AB'$ and $AB$, respectively, and substitute them into the Clauser-Horne-Shimony-Holt variant of Bell's inequality (Clauser et al., 1969)
\begin{equation} \label{chsh}
-2 \le E(A',B')+E(A',B)+E(A,B')-E(A,B) \le 2.
\end{equation}
We performed an experiment involving 81 participants who were presented a questionnaire to be filled out in which they were asked to make a choice among the above alternatives in the experiments $AB$, $A'B$, $AB'$ and $A'B'$ for the combination {\it The Animal Acts}. Table 1 contains the results of our experiment.

If we denote by $P(A_1,B_1)$, $P(A_2,B_2)$, $P(A_1,B_2)$, $P(A_2,B_1)$, the probability that {\it The Horse Growls}, {\it The Bear Whinnies},  
{\it The Horse Whinnies}, {\it The Bear Growls}, respectively, is chosen in the coincidence experiment $AB$, and so on, the expectation values are 
\begin{eqnarray}
E(A,B)&=&P(A_1,B_1)+P(A_2,B_2)-P(A_2,B_1)-P(A_1,B_2)=-0.7778  \nonumber \\
E(A',B)&=&P(A'_1,B_1)+P(A'_2,B_2)-P(A'_2,B_1)-P(A'_1,B_2)=0.6543 \nonumber \\
E(A,B')&=&P(A_1,B'_1)+P(A_2,B'_2)-P(A_2,B'_1)-P(A_1,B'_2)=0.3580 \nonumber \\
E(A',B')&=&P(A'_1,B'_1)+P(A'_2,B'_2)-P(A'_2,B'_1)-P(A'_1,B'_2)=  0.6296 \nonumber
\end{eqnarray}
Hence, the Clauser-Horne-Shimony-Holt variant of Bell's inequalities gives 
\begin{equation}
E(A',B')+E(A',B)+E(A,B')-E(A,B)=2.4197
\end{equation}
which is greater than 2. This means that it violates Bell's inequalities, indeed, it does so close to the maximal possible violation in quantum theory, viz. $2\cdot\sqrt{2} \approx 2.8284$.
\begin{table}[H] 
\begin{footnotesize}
\begin{center}
\begin{tabular}{|c |c | c | c| c| }
\hline
\textrm{$AB$} & \emph{Horse Growls} & \emph{Horse Whinnies} & \emph{Bear Growls} & \emph{Bear Whinnies}\\
 & $P(A_1,B_1)=0.049$ & $P(A_1,B_2)=0.630$ & $P(A_2,B_1)=0.259$ & $P(A_2,B_2)=0.062$  \\
\hline
\hline
\textrm{$A'B$} & \emph{Tiger Growls} & \emph{Tiger Whinnies} & \emph{Cat Growls} & \emph{Cat Whinnies}\\
 & $P(A'_1,B_1)=0.778$ & $P(A'_1, B_2)=0.086$ & $P(A'_2,B_1)=0.086$  &  $P(A'_2,B_2)=0.049$ \\
\hline
\hline
\textrm{$AB'$} & \emph{Horse Snorts} & \emph{Horse Meows} & \emph{Bear Snorts} & \emph{Bear Meows}\\
& $P(A_1,B'_1)=0.593$ & $P(A_1, B'_2)=0.025$ & $P(A_2,B'_1)=0.296$   & $P(A_2,B'_2)=0.086$ \\
\hline
\hline
\textrm{$A'B'$} & \emph{Tiger Snorts} & \emph{Tiger Meows} & \emph{Cat Snorts} & \emph{Cat Meows}\\
 & $P(A'_1,B'_1)=0.148$ & $P(A'_1, B'_2)=0.086$ & $P(A'_2,B'_1)=0.099$ & $P(A'_2,B'_2)=0.667$\\
\hline
\end{tabular}
\end{center}
\end{footnotesize}
\caption{The data collected with our experiment on entanglement in concepts.}
\end{table}
\vspace{-0.2cm}
\noindent
The violation we detected is very significant in proving the presence of entanglement between the concept {\it Animal} and the concept {\it Acts} in the combination {\it The Animal Acts}, because it can be shown that effects of disturbance of the experiment push the value of the Bell expression in equation (\ref{chsh}) towards a value between -2 and +2.
In Aerts \& Sozzo (2011) we performed a statistical analysis of the empirical data using the `t-test for paired two samples for means' to estimate the probability that the shifts from Bell's inequalities is due to chance. We compared the data collected in the real experiment with the data collected in the `classical' experiment, where no influence of context and meaning is present. It was possible to conclude convincingly that the deviation effects were not caused by random fluctuations. For a further discussion of different aspect related to this results, we refer to Aerts \& Sozzo (2011).

We note that the fundamental role played by entanglement in concept combination and word association was pointed out by Nelson and McEvoy (2007) and Bruza et al. (2008, 2009, 2011). It was shown that if one assumes that words can become entangled in the human mental lexicon, then one can provide a unified theoretical framework in which two seemingly competing approaches for modeling the activation level of words in human memory, namely, the `Spreading Activation' and the `Spooky-activation-at-a-distance', can be recovered. 

\vspace{-0.4cm}
\section{Interference \label{interferencesuperposition}}
\vspace{-0.4cm}
We already brought up interference at length in section \ref{inklings}, with respect to the guppy effect and more concretely the membership experiments of Hampton (1988a). In this section, we illustrate the phenomenon of interference for concept combinations in a way that is closer to its essence than in the case of membership weights, where it appeared historically for the first time as a consequence of the Hampton (1988a) measurements. In section \ref{glimmers}, we mentioned how we introduced the notion of state, and the notion of change of state under the influence of a context, and how exemplars of a concept in our SCoP approach are states of this concept. Change of state is often called collapse in quantum theory, more specifically when it is a change of state provoked by an experiment or a decision process. The state before the experiment or decision is said to collapse to one of the possible states after the experiment or decision. If we consider our example of the {\it The Animal Acts}, and we denote the state before the experiment or decision by $p_{Animal\ Acts}$, then this state changes to one of the exemplar states $p_{Horse\ Growls}$, $p_{Horse\ Whinnies}$, $p_{Bear\ Growls}$ or $p_{Bear\ Whinnies}$. Hence, this collapse is a change from a more abstract state to more concrete states, expressed by exemplars of the original concept. Also, with regard to interference taking place with combining concepts, we want to look at it for quantum collapses taking place from abstract concept states to more concrete exemplar states.

Let us consider the two concepts {\it Fruits} and {\it Vegetables}, and their disjunction {\it Fruits or Vegetables}. We consider a collection of exemplars, more specifically those listed in Table 2, which are the ones Hampton (1988b) experimented on for the disjunction of {\it Fruits} and {\it Vegetables}. Then we consider the experiment consisting in participants being asked to choose one of the exemplars from the list of Table 2 that they find `a good example of' {\it A = Fruits}, {\it B = Vegetables}, and {\it A or B = Fruits or Vegetables}, respectively. The quantities $\mu(A)_k$, $\mu(B)_k$ and $\mu(A\ {\rm or}\ B)_k$ are the relative frequencies of the outcomes for this experiment, which are also given in Table 2. Let us remark that Hampton (1988b) did not perform this experiment, since he was testing the membership weights for the concepts and their disjunction. However, in his experiment, Hampton (1988b) tested for the `degree of membership' by asking the participants, 40 undergraduate psychology students, to rate the membership on a Likert scale $\{-3, -2, -1, 0, +1, +2, +3\}$, where positive numbers stand for `membership', and negative numbers for `non-membership', the size of the number corresponding to `the degree of membership or non-membership'. In view of the very strong correlation between `degree of membership' and `a good example of', we used the data of Hampton (1988b) to calculate the relative frequencies. In future research, we intend to conduct an experiment directly measuring `good example of', in a similar manner as in our study of entanglement and {\it The Animal Acts}.

We constructed an explicit quantum mechanical model in complex Hilbert space for the pair of concepts {\it Fruit} and {\it Vegetable} and their disjunction {\it Fruit or Vegetable}, with respect to the relative frequencies calculated from Hampton (1988b) experimental data (Aerts, 2009b, section 3). We represent here its main aspects following the quantum modeling scheme explained in detail in appendix A. The measurement `a good example of' is represented by a spectral family $\{M_k\ \vert\ k=1,\ldots,24\}$, where each $M_k$ is an orthogonal projection of the Hilbert space ${\mathbb C}^{25}$ corresponding to item $k$ from the list of items in Table 2. 

The concepts {\it Fruits}, {\it Vegetables} and {\it Fruits or Vegetables} are represented by unit vectors $|A\rangle$, $|B\rangle$ and ${1 \over \sqrt{2}}(|A\rangle+|B\rangle)$ of this Hilbert space, where $|A\rangle$ and $|B\rangle$ are orthogonal, and ${1 \over \sqrt{2}}(|A\rangle+|B\rangle)$ is their normalized superposition. Following standard quantum rules, we have $\mu(A)_k=\langle A|M_k|A\rangle$, $\mu(B)_k=\langle B|M_k|B\rangle$, and
\begin{eqnarray} 
&\!\!\!\! \mu(A\ \! {\rm or}\ \! B)_k \! = \! ({1 \over \sqrt{2}}(\langle A| \! + \! \langle B|)M_k({1 \over \sqrt{2}}(|A\rangle \! + \! |B\rangle)) \! = \! {1 \over 2}(\langle A|M_k|A\rangle \! + \! \langle B|M_k|B\rangle \! + \! \langle A|M_k|B\rangle \! + \! \langle B|M_k|A\rangle) \nonumber \\ 
& \!\!\!\! \!\!\!\! \!\!\!\! \!\!\!\! \!\!\!\! \!\!\!\! \!\!\!\! \!\!\!\!  \!\!\!\! \!\!\!\! \!\!\!\! \!\!\!\! \!\!\!\! \!\!\!\! \!\!\!\! \!\!\!\!  \!\!\!\! \!\!\!\! \!\!\!\! \!\!\!\! \!\!\!\! \!\!\!\! \!\!\!\! \!\!\!\! \!\! \!\!\!\! \!\!\!\! ={1 \over 2}(\mu(A)_k+\mu(B)_k)+\Re\langle A|M_k|B\rangle \label{muAorB}
\end{eqnarray}
where $\Re\langle A|M_k|B\rangle$, the real part of the complex number $\langle A|M_k|B\rangle$, is the interference term.

Remark the similarity with equation \ref{membershipweightinterference} of section \ref{inklings}. We showed in Aerts (2009b) that we can find a solution for
\begin{equation} \label{muAorBequation}
\Re\langle A|M_k|B\rangle=c_k\sqrt{\mu(A)_k\mu(B)_k}\cos\phi_k \qquad \cos\phi_k={2\mu(A\ {\rm or}\ B)_k-\mu(A)_k-\mu(B)_k \over 2c_k\sqrt{\mu(A)_k\mu(B)_k}}
\end{equation}
(see Table 2), where $c_k$ is chosen in such a way that $\langle A|B\rangle=0$.
\begin{table}[H]
\small
\begin{center}
\begin{tabular}{|lllllll|}
\hline 
\multicolumn{2}{|l}{} & \multicolumn{1}{l}{$\mu(A)_k$} & \multicolumn{1}{l}{$\mu(B)_k$} & \multicolumn{1}{l}{$\mu(A\ {\rm or}\ B)_k$}
& \multicolumn{1}{l}{${1 \over 2}(\mu(A)_k+\mu(B)_k)$} & \multicolumn{1}{l|}{$\phi_k$} \\
\hline
\multicolumn{7}{|l|}{\it $A$=Fruits, $B$=Vegetables} \\
\hline
1 & {\it Almond} & 0.0359 & 0.0133 & 0.0269 & 0.0246 & 83.8854$^\circ$ \\
2 & {\it Acorn} & 0.0425 & 0.0108 & 0.0249 & 0.0266 &  -94.5520$^\circ$ \\
3 & {\it Peanut} & 0.0372 & 0.0220 & 0.0269 & 0.0296 &  -95.3620$^\circ$ \\
4 & {\it Olive} & 0.0586 & 0.0269 & 0.0415 & 0.0428 &  91.8715$^\circ$ \\
5 & {\it Coconut} & 0.0755 & 0.0125 & 0.0604 & 0.0440 & 57.9533$^\circ$ \\
6 & {\it Raisin} & 0.1026 & 0.0170 & 0.0555 & 0.0598 &  95.8648$^\circ$ \\
7 & {\it Elderberry} & 0.1138 & 0.0170 & 0.0480 & 0.0654 & -113.2431$^\circ$ \\ 
8 & {\it Apple} & 0.1184 & 0.0155 & 0.0688 & 0.0670 &  87.6039$^\circ$ \\ 
9 & {\it Mustard} & 0.0149 & 0.0250 & 0.0146 & 0.0199 &  -105.9806$^\circ$ \\
10 & {\it Wheat} & 0.0136 & 0.0255 & 0.0165 & 0.0195 &  99.3810$^\circ$ \\ 
11 & {\it Root Ginger} & 0.0157 & 0.0323 & 0.0385 & 0.0240 &  50.0889$^\circ$ \\
12 & {\it Chili Pepper} & 0.0167 & 0.0446 & 0.0323 & 0.0306 &   -86.4374$^\circ$ \\ 
13 & {\it Garlic} & 0.0100 & 0.0301 & 0.0293 & 0.0200 & -57.6399$^\circ$ \\
14 & {\it Mushroom} & 0.0140 & 0.0545 & 0.0604 & 0.0342 &  18.6744$^\circ$ \\
15 & {\it Watercress} & 0.0112 & 0.0658 & 0.0482 & 0.0385 &  -69.0705$^\circ$ \\
16 & {\it Lentils} & 0.0095 & 0.0713 & 0.0338 & 0.0404 & 104.7126$^\circ$ \\
17 & {\it Green Pepper} & 0.0324 & 0.0788 & 0.0506 & 0.0556 &  -95.6518$^\circ$ \\
18 & {\it Yam} & 0.0533 & 0.0724 & 0.0541 & 0.0628 &  98.0833$^\circ$ \\
19 & {\it Tomato} & 0.0881 & 0.0679 & 0.0688 & 0.0780 & 100.7557$^\circ$ \\
20 & {\it Pumpkin} & 0.0797 & 0.0713 & 0.0579 & 0.0755 & -103.4804$^\circ$  \\
21 & {\it Broccoli} & 0.0143 & 0.1284 & 0.0642 & 0.0713 & -99.6048$^\circ$ \\
22 & {\it Rice} & 0.0140 & 0.0412 & 0.0248 & 0.0276 & -96.6635$^\circ$ \\ 
23 & {\it Parsley} & 0.0155 & 0.0266 & 0.0308 & 0.0210 & -61.1698$^\circ$ \\
24 & {\it Black Pepper} & 0.0127 & 0.0294 & 0.0222 & 0.0211 &  86.6308$^\circ$ \\
\hline
\end{tabular}
\end{center}
\caption{Interference data for concepts {\it A=Fruits} and {\it B=Vegetables}. The probability of being chosen as `a good example of' {\it Fruits} ({\it Vegetables}) is $\mu(A)$ ($\mu(B)$) for each of the exemplars. The probability of being chosen as `a good example of' {\it Fruits or Vegetables} is $\mu(A\ {\rm or}\ B)$ for each of the exemplars. The classical probability is ${\mu(A)+\mu(B) \over 2}$ and $\theta$ is the quantum interference angle.
}
\end{table}
\normalsize
\noindent
We construct an explicit solution with the following vectors 
\begin{eqnarray} \label{interferenceangles01}
|A\rangle=(0.1895, 0.2061, 0.1929, 0.2421, 0.2748, 0.3204, 0.3373, 0.3441, 
           0.1222, 0.1165, 0.1252, 0.1291, \nonumber \\  0.1002, 0.1182, 0.1059, 0.0974, 0.1800,    
           0.2308, 0.2967, 0.2823, 0.1194, 0.1181, 0.1245, 0.1128, 0) \nonumber
\end{eqnarray}
\begin{eqnarray} 
|B\rangle=(0.1154e^{i83.8854^\circ}, 0.1040e^{-i94.5520^\circ}, 0.1484e^{-i95.3620^\circ}, 0.1640e^{i91.8715^\circ}, 
           0.1120e^{i57.9533^\circ},  \nonumber \\ 0.1302e^{i95.8648^\circ},  
           0.1302e^{-i113.2431^\circ},  0.1246e^{i87.6039^\circ},  
           0.1580e^{-i105.9806^\circ}, 0.1596e^{i99.3810^\circ}, \nonumber \\ 
           0.1798e^{i50.0889^\circ},  0.2112e^{-i86.4374^\circ},  
           0.1734e^{-i57.6399^\circ},  0.2334e^{i18.6744^\circ}, 0.2565e^{-i69.0705^\circ}, \nonumber \\ 0.2670e^{i104.7126^\circ}, 
           0.2806e^{-i95.6518^\circ},  0.2690e^{i98.0833^\circ},   0.2606e^{i100.7557^\circ}, 0.2670e^{-i103.4804^\circ}, \nonumber \\
           0.3584e^{-i99.6048^\circ}, 0.2031e^{-i96.6635^\circ},  0.1630e^{-i61.1698^\circ}, 0.1716e^{i86.6308^\circ},  0.1565) \label{interferenceangles02}
\end{eqnarray}
We worked out a way to represent graphically the quantum interference of {\it Fruits} with {\it Vegetables} (Aerts, 2009b). For the concepts {\it Fruits}, {\it Vegetables} and {\it Fruits or Vegetables} we use complex valued wave functions of two real variables $\psi_A(x,y), \psi_B(x,y)$ and $\psi_{A{\rm or}B}(x,y)$ to represent them. We choose $\psi_A(x,y)$ and $\psi_B(x,y)$ such that the real part for both wave functions is a Gaussian in two dimensions, such that the top of the first Gaussian is in the origin, and the top of the second Gaussian is located in the point $(a,b)$. Hence
\begin{equation}
\psi_A(x,y)=\sqrt{D_A}e^{-({x^2 \over 4\sigma^2_{Ax}}+{y^2 \over 4\sigma^2_{Ay}})}e^{iS_A(x,y)} \qquad \psi_B(x,y)=\sqrt{D_B}e^{-({(x-a)^2 \over 4\sigma^2_{Bx}}+{(y-b)^2 \over 4\sigma^2_{By}})}e^{iS_B(x,y)}
\end{equation}
The squares of these Gaussians -- also Gaussians -- represent the membership weights and are graphically represented in Figures 1 and 2, and the different exemplars of Table 2 are located in spots such that the Gaussian distributions $|\psi_A(x,y)|^2$ and $|\psi_B(x,y)|^2$ properly model the probabilities $\mu(A)_k$ and $\mu(B)_k$ in Table 2 for each one of the exemplars.
\begin{figure}[H]
\centerline {\includegraphics[scale=0.52]{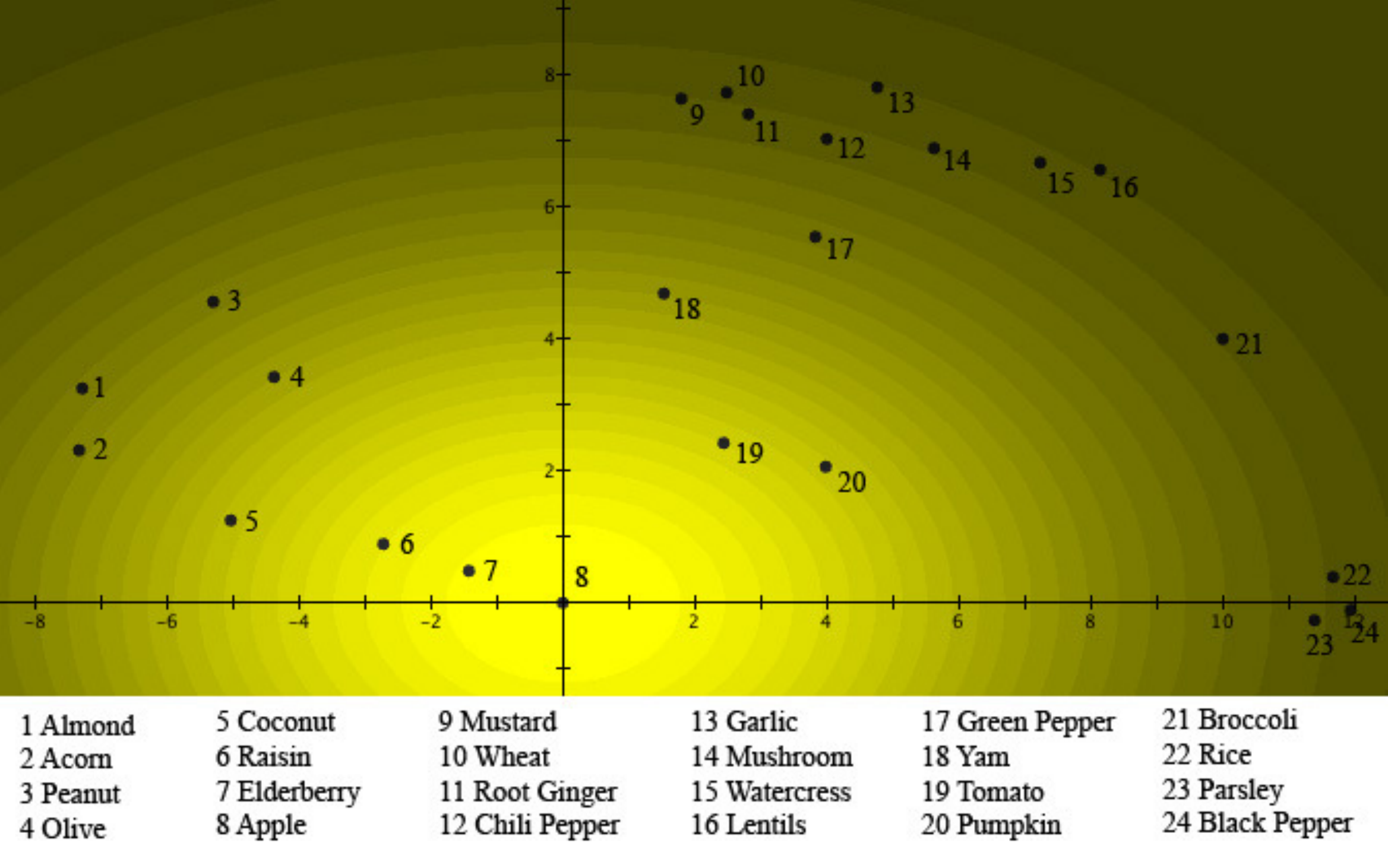}}
\caption{The probabilities $\mu(A)_k$ of a person choosing the exemplar $k$ as a `good example of' {\it Fruits} are fitted into a two-dimensional quantum wave function $\psi_A(x,y)$. The numbers are placed at the locations of the different exemplars with respect to the Gaussian probability distribution $|\psi_A(x,y)|^2$. This can be seen as a light source shining through a hole centered on the origin, and regions where the different exemplars are located. The brightness of the light source in a specific region corresponds to the probability that this exemplar will be chosen as a `good example of' {\it Fruits}.
}
\end{figure}
\noindent
This is always possible, taking into account the parameters of the Gaussians, $D_A$, $D_B$, $\sigma_{Ax}$, $\sigma_{Ay}$, $\sigma_{Bx}$, $\sigma_{By}$, $a$, $b$ and the necessity to fit 24 values, namely the values of $\mu(A)_k$ and $\mu(B)_k$ for each of the exemplars of Table 2. Remark that the constraint comes from the exemplars having to be located in exactly the same points of the plane for both Gaussians. Although the solution is an elaborate mathematical calculation, it is also straightforward, so that we leave its elaboration to the interested and exploring reader. It is not unique, but different solutions are topologically stretched versions of the one we the one we use in this article, which means that the interference pattern of other solutions is topomorphic to the one we present here.
\begin{figure}[H]
\centerline {\includegraphics[scale=0.52]{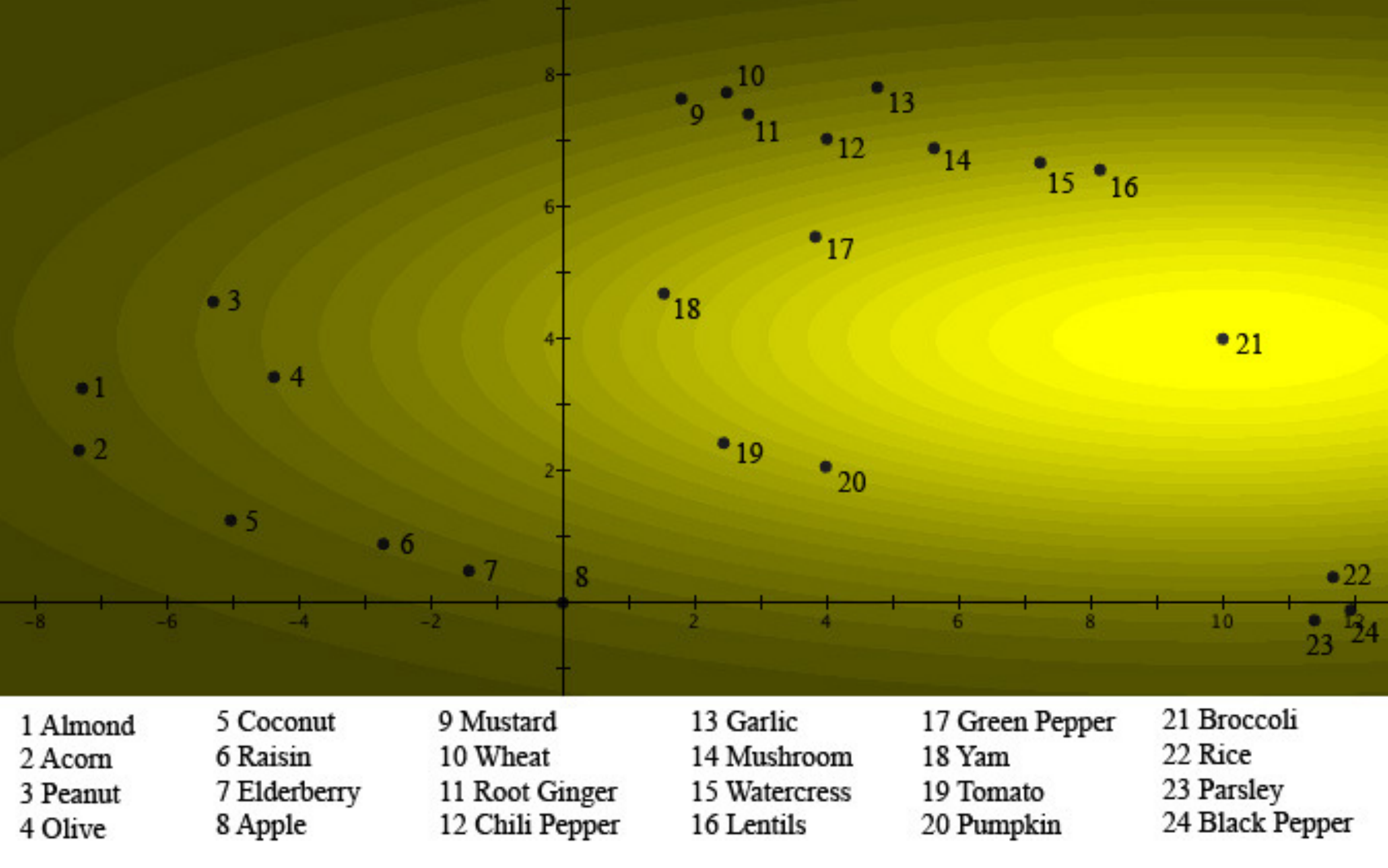}}
\caption{The probabilities $\mu(B)_k$ of a person choosing the exemplar $k$ as a `good example of' {\it Vegetables} are fitted into a two-dimensional quantum wave function $\psi_B(x,y)$. The numbers are placed at the locations of the different exemplars with respect to the probability distribution $|\psi_B(x,y)|^2$. This can be seen as a light source shining through a hole centered on point 21, where {\it Broccoli} is located. The brightness of the light source in a specific region corresponds to the probability that this exemplar will be chosen as a `good example of' {\it Vegetables}.
}
\end{figure}
\begin{figure}[H]
\centerline {\includegraphics[scale=0.52]{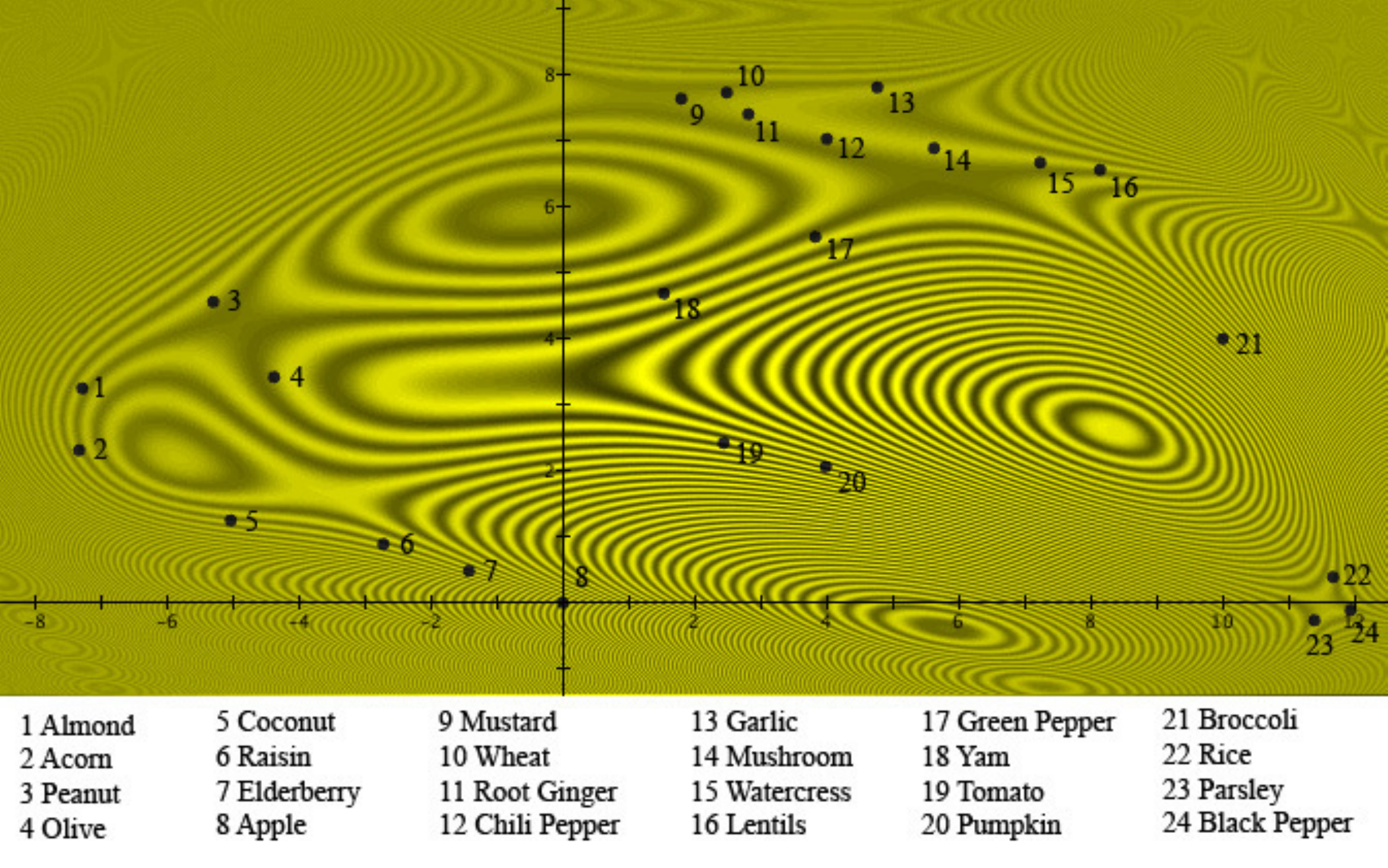}}
\caption{The probabilities $\mu(A\ {\rm or}\ B)_k$ of a person choosing the exemplar $k$ as a `good example of' `{\it Fruits or Vegetables}' are fitted into the two-dimensional quantum wave function ${1 \over \sqrt{2}}(\psi_A(x,y)+\psi_B(x,y))$, which is the normalized superposition of the wave functions in Figures 1 and 2. The numbers are placed at the locations of the different exemplars with respect to the probability distribution ${1 \over 2}|\psi_A(x,y)+\psi_B(x,y)|^2={1 \over 2}(|\psi_A(x,y)|^2+|\psi_B(x,y)|^2)+|\psi_A(x,y)\psi_B(x,y)|\cos\phi(x,y)$, where $\phi(x,y)$ is the quantum phase difference at $(x,y)$. The values of $\phi(x,y)$ are given in Table 1 for the locations of the different exemplars. The interference pattern is clearly visible.
}
\end{figure}
\noindent
For {\it Fruits} represented in Figure 1, {\it Apple} is located in the center of the Gaussian, since {\it Apple} appears most as a `good example of' {\it Fruits}. {\it Elderberry} is second, and hence closest to the top of the Gaussian in Figure 1. Then come {\it Raisin}, {\it Tomato} and {\it Pumpkin}, and so on, with {\it Garlic} and {\it Lentils} the least chosen `good examples' of {\it Fruits}. For {\it Vegetables}, represented in Figure 2, {\it Broccoli} is located in the center of the Gaussian, since {\it Broccoli} appears most as a `good example of' {\it Vegetables}. {\it Green Pepper} is second, and hence closest to the top of the Gaussian in Figure 2. Then come {\it Yam}, {\it Lentils} and {\it Pumpkin}, and so on, with {\it Coconut} and {\it Acorn} as the least chosen `good examples' of {\it Vegetables}. Metaphorically, we can regard the graphical representations of Figures 1 and 2 as the projections of a light source shining through one of two holes in a plate and spreading out its light intensity following a Gaussian distribution when projected on a screen behind the holes.

The center of the first {\it Fruits} hole is located where exemplar {\it Apple} is at point $(0, 0)$, indicated by 8 in both figures. The center of the second {\it Vegetables} hole is located where exemplar {\it Broccoli} is at point (10,4), indicated by 21 in both figures.

In Figure 3, the data for {\it Fruits or Vegetables} are graphically represented. This is the probability distribution corresponding to ${1 \over \sqrt{2}}(\psi_A(x,y)+\psi_B(x,y))$, which is the normalized superposition of the wave functions in Figures 1 and 2, which is not `just' a normalized sum of the two Gaussians of Figures 1 and 2. The numbers are placed at the locations of the different exemplars with respect to the probability distribution ${1 \over 2}|\psi_A(x,y)+\psi_B(x,y)|^2={1 \over 2}(|\psi_A(x,y)|^2+|\psi_B(x,y)|^2)+|\psi_A(x,y)\psi_B(x,y)|\cos\phi(x,y)$, where $|\psi_A(x,y)$$\psi_B(x,y)|\cos\phi(x,y)$ is the interference term and $\phi(x,y)=S_A(x,y)-S_B(x,y)$ the quantum phase difference at $(x,y)$. We calculated a solution for the function $\phi(x,y)$ as a linear combination of powers of products of $x$ and $y$, electing the lowest powers to attain 25 linear independent functions, with the constraint that the values of $\phi(x,y)$ coincide with the ones in Table 2 for the locations of the different exemplars. It can be proven that such a solution exists and is unique.

The interference pattern shown in Figure 3 is very similar to well-known interference patterns of light passing through an elastic material under stress. In our case, it is the interference pattern corresponding to {\it Fruits or Vegetables}. Bearing in mind the analogy with the light source for Figures 1 and 2, in Figure 3 we can see the interference pattern produced when both holes are open. Figure 4 represents a three-dimensional graphic of the interference pattern of Figure 3, and, for the sake of comparison, in Figure 5, we have graphically represented the averages of the probabilities of Figures 1 and 2, i.e. the values measured if there were no interference. For a more detailed analysis we refer to Aerts (2009b), and we put forward the graphical representation of an example of conceptual interference for the conjunction of two concepts in Aerts et al (2012).

The expression of vector $|B\rangle$ in equation (\ref{interferenceangles02}) confirms the underground nature of quantum theory in a convincing way. We see that each of the 25 components is intrinsically complex with different interference angles. Indeed, it can be proven that a representation in a real vectors space cannot reproduce the interference pattern for {\it Fruits or Vegetables}, as graphically shown in Figures 3 and 4. 

A very interesting similarity can be noted between the type of modeling we explore in this section, and semantic space approaches in computer science, where terms and documents are represented in a term-document matrix, which makes it possible to introduce a real vector space model (Aerts \& Czachor, 2004). Latent Semantic Analysis is a well-known and powerful example of these approaches (Deerwester et al., 1990).
\begin{figure}[H]
\centerline {\includegraphics[scale=0.44]{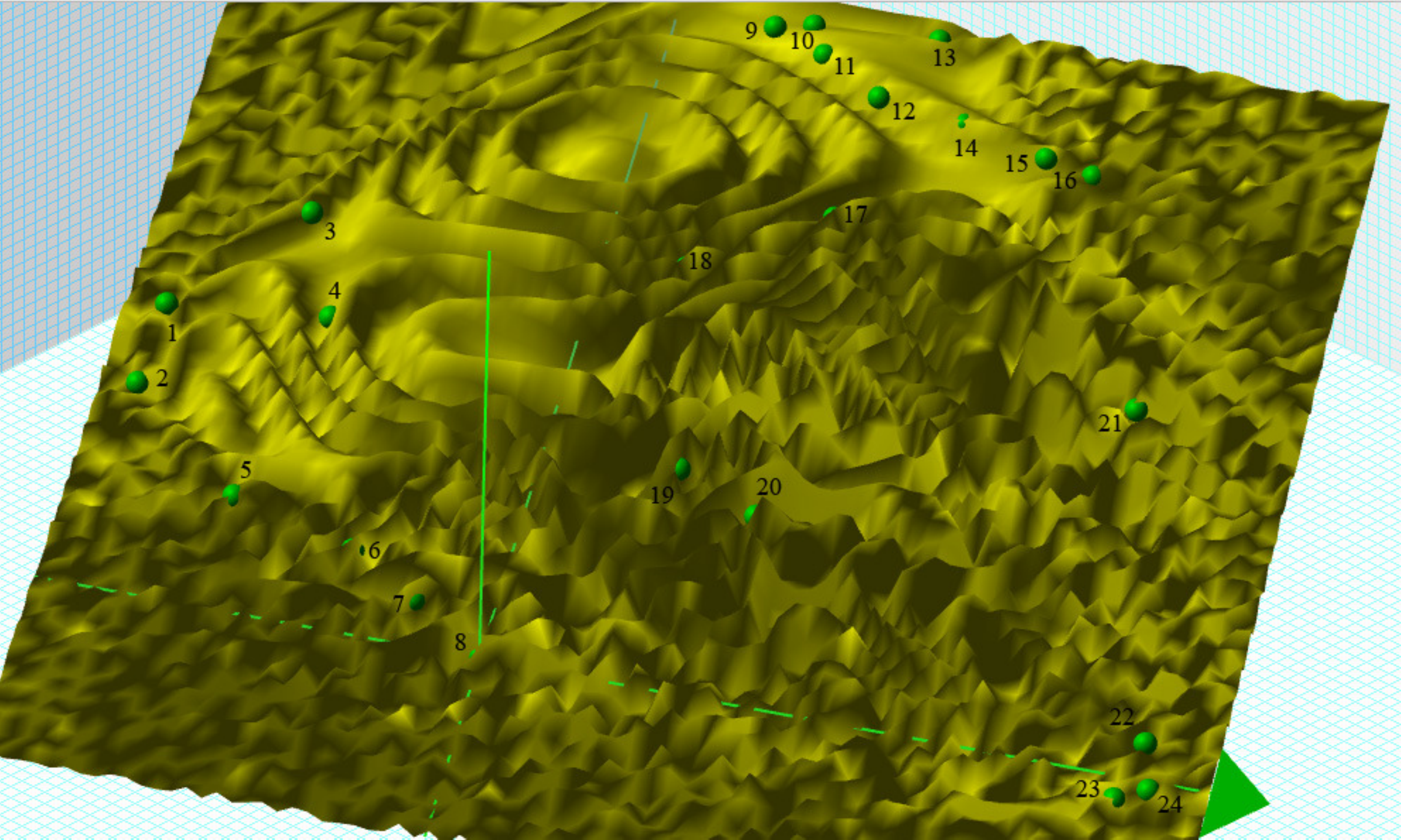}}
\caption{A three-dimensional representation of the interference landscape of the concept `{\it Fruits {\rm or} Vegetables}' as shown in Figure 3. Exemplars are represented by little green balls, and the numbers refer to the numbering of the exemplars in Table 1 and in Figures 1, 2 and 3.
}
\end{figure}
\begin{figure}[H]
\centerline {\includegraphics[scale=0.52]{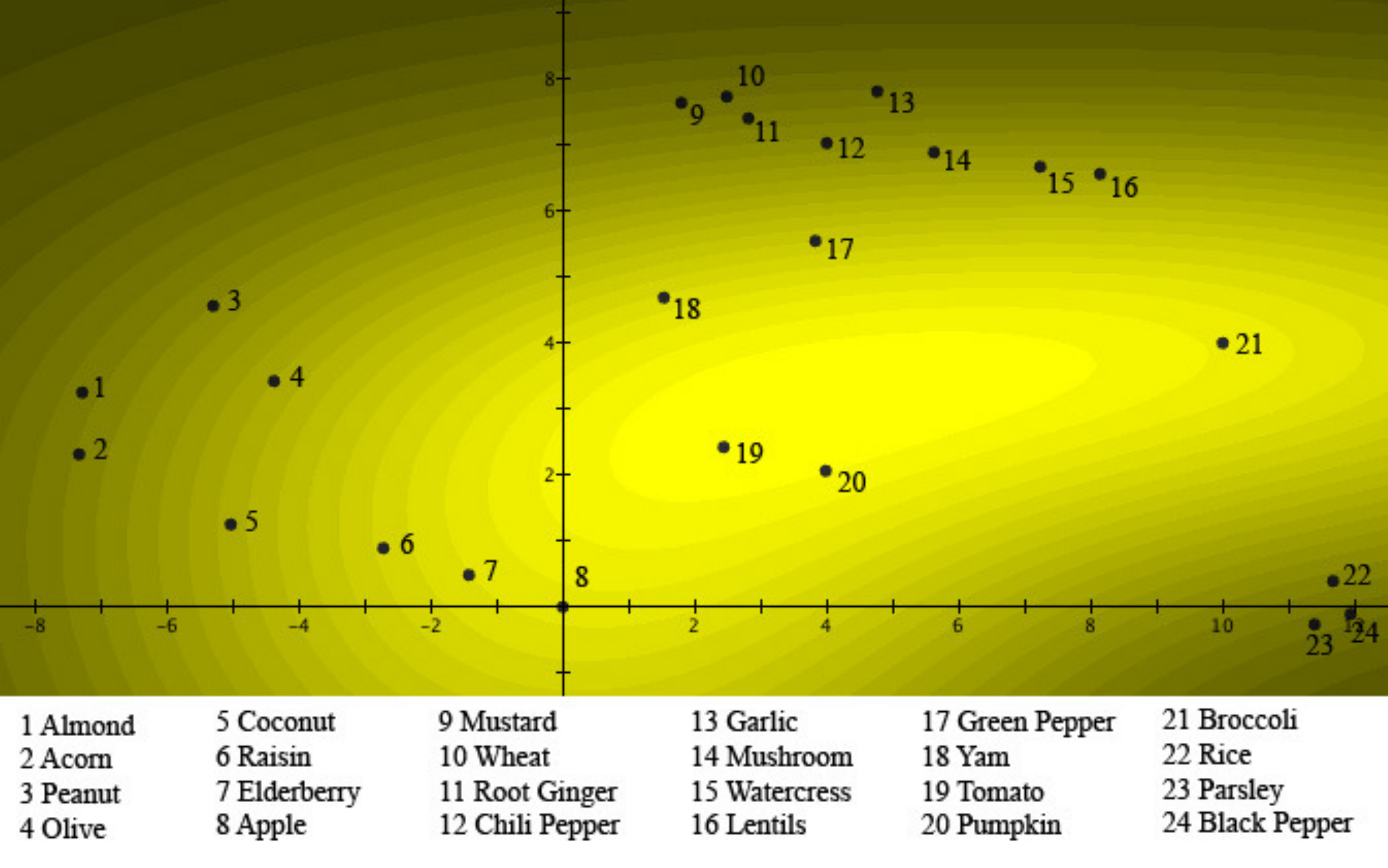}}
\caption{Probabilities $1/2(\mu(A)_k+\mu(B)_k)$, which are the probability averages for {\it Fruits} and {\it Vegetables} shown in Figures 1 and 2. This would be the resulting pattern in case $\phi(x,y)=90^\circ$ for all exemplars. It is called the classical pattern for the situation since it is the pattern that, without interference, results from a situation where classical particles are sent through two slits. These classical values for all exemplars are given in Table 1.
}
\end{figure}
\noindent
But connections with quantum structures have been indicated also for information retrieval, an exploding and important domain in computer science building on semantic space approaches (Van Rijsbergen, 2004; Widdows \& Peters, 2003) and they are now explored intensively (Li \& Cunningham, 2008; Melucci, 2008; Piwowarski et al., 2010; Widdows, 2006; Zuccon \& Azzopardi, 2010). Our guess is that the structure of the real -- and not complex -- vector space that governs the traditional semantic spaces, and the derived information retrieval schemes, will turn out to correspond to a provisional stage on the way toward a `complex number semantic space scheme'. Hence it will be a fascinating challenge further to increase our understanding  of the role of complex numbers, also with respect to how `terms' -- in our view `concepts' -- relate to `documents' -- in our view `conceptual contexts'. We hope that the type of intrinsically complex model we put forward in this section can help shed light on this aspect.

\vspace{-0.4cm}
\section{Emergence and Potentiality\label{emergence}}
\vspace{-0.4cm}
The guppy effect and the effects of overextension and underextension identified in concept research by Osherson (1981) and Hampton (1988a,b), respectively, have their counterparts in other domains of psychology. There is a whole set of findings, often referred to as originating from the `Tversky and Khaneman program', and focusing mainly on human decision making, that entail effects of a very similar nature, e.g. the disjunction effect and the conjunction fallacy (Tversky \& Khaneman, 1983; Tversky \& Shafir, 1992). In economics, similar effects have been found and identified that point to a deviation from classical logical thinking when human decisions are at stake, even before this happened in psychology (Allais, 1953; Ellsberg, 1962). 
In addition, decision researchers have discovered the value of quantum modeling, making fruitful use of quantum decision models for the modeling of a large number of experimentally identified effects (Busemeyer et al., 2006, 2011; Lambert Mogiliansky et al., 2009; Pothos \& Busemeyer, 2009).

The tendency within the domain of decision-making, certainly in the early years, was to consider these deviations from classicality as `fallacies' -- cf. the conjunction fallacy -- or, if not fallacies, as `effects' -- cf. the disjunction effect. Concept researchers too imagined that an effect was at play, i.e. the guppy effect. We ourselves have referred to the phenomenon using such terms and phrases as `effect' and `deviation from classicality'. In this section, we want to show that the state of affairs is in fact the other way around. What has been called a fallacy, an effect or a deviation, is a consequence of the dominant dynamics, while what has been considered as a default to deviate from, namely classical logical reasoning, is a consequence of a secondary form of dynamics. The dominant dynamics does not give rise to classical logical reasoning. So what is the nature of this dominant dynamics? Its nature is emergence. 

Let us put forward the argumentation that underlies the above hypothesis and that we have arrived at by analyzing the experiments by Hampton (1988a,b) and our experiments on entanglement (Aerts \& Sozzo, 2011). In Table 3 we have presented the exemplars and pairs we want to consider for our argumentation. A complete collection of all the membership weight data can be found in Tables 4 and 3 in Aerts (2009a).

An important role is played by the abundance of exemplars with overextension in case of conjunction, and with underextension in case of disjunction, except for the pair {\it Fruits} and {\it Vegetables}. This is the one we modeled in section \ref{interferencesuperposition}, where disjunction gives rise to overextension too, and in such a strong way, that we will start from there.

So let us consider {\it Mushroom}, whose membership weight for {\it Fruits or Vegetables} is 0.9, while its membership weight for {\it Fruits} is 0 and for {\it Vegetables} 0.5 (Table 3). 
\begin{table}[H]
\footnotesize
\begin{center}
\begin{tabular}{|llllllll|}
\hline
\multicolumn{2}{|l}{} & \multicolumn{1}{l}{$\mu(A)$} & \multicolumn{1}{l}{$\mu(B)$} & \multicolumn{1}{l}{$\mu(A\ {\rm or}\ B)$} & \multicolumn{1}{l}{$\Delta_d$} & \multicolumn{1}{l}{$k_d$} & \multicolumn{1}{l|}{$f_d$} \\
\hline
\multicolumn{8}{|l|}{\it $A$=Fruits, $B$=Vegetables} \\
\hline
k & {\it Mushroom} & 0 & 0.5 & 0.9 & -0.4 & -0.4 & -0.4 \\
k & {\it Parsley} & 0 & 0.2 & 0.45 & -0.25 & -0.25 & -0.25 \\
k & {\it Olive} & 0.5 & 0.1 & 0.8 & -0.3 & -0.2 & -0.25 \\
k & {\it Root Ginger} & 0 & 0.3 & 0.55 & -0.25 & -0.25 & -0.25 \\
k & {\it Acorn} & 0.35 & 0 & 0.4 & -0.05 & -0.05 & -0.05 \\
k & {\it Garlic} & 0.1 & 0.2 & 0.5 & -0.3 & -0.2 & -0.22 \\
k & {\it Almond} & 0.2 & 0.1 & 0.43 & -0.23 & -0.13 & -0.15 \\
c & {\it Tomato} & 0.7 & 0.7 & 1 & -0.3 & 0.4 & -0.09 \\
c & {\it Pumpkin} & 0.7 & 0.8 & 0.93 & -0.13 & 0.58 & 0.02 \\
$\Delta$ & {\it Mustard} & 0 & 0.2 & 0.18 & 0.03 & 0.03 & 0.03 \\
\hline
\multicolumn{8}{|l|}{\it $A$=Home Furnishings, $B$=Furniture} \\
\hline
$\Delta$ & {\it Ashtray} & 0.3 & 0.7 & 0.25 & 0.45 & 0.75 & -0.25 \\
$\Delta$ & {\it Painting} & 0.5 & 0.9 & 0.85 & 0.05 & 0.55 & 0.1 \\
c & {\it Deck Chair} & 0.3 & 0.1 & 0.35 & -0.05 & 0.05 & 0.02 \\
\hline
\multicolumn{8}{|l|}{\it $A$=Hobbies, $B$=Games} \\
\hline
$\Delta$ & {\it Discus Throwing} & 1 & 0.75 & 0.7 & 0.3 & 1.05 & -0.18 \\
$\Delta$ & {\it Camping} & 1 & 0.1 & 0.9 & 0.1 & 0.2 & 0.1 \\
c & {\it Gardening} & 1 & 0 & 1 & 0 & 0 & 0 \\
\hline
\multicolumn{8}{|l|}{\it $A$=Instruments, $B$=Tools} \\
\hline
$\Delta$ & {\it Bicycle Pump} & 1 & 0.9 & 0.7 & 0.3 & 1.2 & -0.25 \\
$\Delta$ & {\it Goggles} & 0.2 & 0.3 & 0.15 & 0.15 & 0.35 & -0.1 \\
c & {\it Tuning Fork} & 0.9 & 0.6 & 1 & -0.1 & 0.5 & -0.04 \\
$\Delta$ & {\it Spoon} & 0.65 & 0.9 & 0.7 & 0.2 & 0.85 & -0.08 \\
k & {\it Door Key} & 0.3 & 0.1 & 0.95 & -0.65 & -0.55 & -0.58 \\
\hline
\multicolumn{8}{|l|}{\it $A$=Pets, $B$=Farmyard Animals} \\
\hline
$\Delta$ & {\it Rat} & 0.5 & 0.7 & 0.4 & 0.3 & 0.8 & -0.2 \\
$\Delta$ & {\it Cart Horse} & 0.4 & 1 & 0.85 & 0.15 & 0.55 & 0.15 \\
\hline
\multicolumn{8}{|l|}{\it $A$=Sportswear, $B$=Sports Equipment} \\
\hline
$\Delta$ & {\it Diving Mask} & 1 & 1 & 0.95 & 0.05 & 1,05 & -0.05 \\
$\Delta$ & {\it Underwater} & 1 & 0.65 & 0,6 & 0.4 & 1.05 & -0.23 \\
\hline
\hline
\multicolumn{2}{|l}{} & \multicolumn{1}{l}{$\mu(A)$} & \multicolumn{1}{l}{$\mu(B)$} & \multicolumn{1}{l}{$\mu(A\ {\rm and}\ B)$} & \multicolumn{1}{l}{$\Delta_c$} & \multicolumn{1}{l}{$k_c$} & \multicolumn{1}{l|}{$f_c$} \\
\hline
\multicolumn{8}{|l|}{\it $A$=Food, $B$=Plant} \\
\hline
$\Delta$ & {\it Mint} & 0.87 & 0.81 & 0.9 & 0.09 & 0.22 & -0.06 \\
$\Delta$ & {\it Sunflower} & 0.77 & 1 & 0.78 & 0.01 & 0.01 & 0.01 \\
$\Delta$ & {\it Potato} & 1 & 0.74 & 0.9 & 0.16 & 0.16 & -0.03 \\
\hline
\multicolumn{8}{|l|}{\it $A$=Furniture, $B$=Household Appliances} \\
\hline
$\Delta$ & {\it TV} & 0.7 & 0.9 & 0.93 & 0.23 & 0.33 & -0.13 \\
$\Delta$ & {\it Clothes Washer} & 0.15 & 1 & 0.73 & 0.58 & 0.58 & -0.15 \\
$\Delta$ & {\it Hifi} & 0.58 & 0.79 & 0.79 & 0.21 & 0.42 & -0.11 \\
$\Delta$ & {\it Heated Waterbed} & 1 & 0.49 & 0.78 & 0.29 & 0.29 & -0.03 \\
$\Delta$ & {\it Floor Mat} & 0.56 & 0.15 & 0.21 & 0.06 & 0.49 & 0.12 \\
$\Delta$ & {\it Coffee Table} & 1 & 0.15 & 0.38 & 0.23 & 0.23 & 0.19 \\
\hline
\multicolumn{8}{|l|}{\it $A$=Building, $B$=Dwelling} \\
\hline
$\Delta$ & {\it Tree House} & 0.77 & 0.846 & 0.85 & 0.08 & 0.23 & -0.04 \\ 
$\Delta$ & {\it Appartment Block} & 0.92 & 0.87 & 0.92 & 0.051 & 0.13 & -0.03 \\
c & {\it Synagoge} & 0.93 & 0.49 & 0.45 & -0.04 & 0.04 & -0.003 \\ 
c & {\it Phone box} & 0.23 & 0.05 & 0.03 & -0.02 & 0.74 & 0.02 \\
\hline
\multicolumn{8}{|l|}{\it $A$=Machine, $B$=Vehicle} \\
\hline
$\Delta$ & {\it Course Liner} & 0.88 & 0.88 & 0.95 & 0.08 & 0.2 &  -0.08 \\ 
\hline
\end{tabular}
\end{center}
\caption{Membership weight data in Hampton (1988a,b)
}
\end{table}
\normalsize
\noindent
This means that participants estimated {\it Mushroom} to be a much stronger member of {\it Fruits or Vegetables} than of {\it Fruits} and {\it Vegetables} apart. This overextension is so extremely strong, however, that we will consider it more in detail. None of the participants found {\it Mushroom} to be a member of {\it Fruits}, and only half of them found it to be a member of {\it Vegetables}, while $90\%$ found it to be a member of {\it Fruits or Vegetables}. This means that at least $40\%$ of the participants found {\it Mushroom} to be `not a member of {\it Fruits}', and `not a member of {\it Vegetables}', but did find it to be `a member of {\it Fruits or Vegetables}'. This defies even the wildest interpretation of a classical logical structure for the disjunction. Additionally, in Aerts (2009a) we proved that the experimental data for {\it Mushroom} cannot be fitted into any type of classical probability structure that could possibly be devised for the disjunction. Indeed, similarly to (\ref{mindeviation}) and (\ref{kolmogorovianfactorconjunction}), two inequalities can be derived for the membership weights and for the disjunction, and only if they are satisfied is a classical probability model possible (Aerts 2009a, theorem 6) 
\begin{eqnarray} \label{maxdeviation}
&&\max(\mu(A),\mu(B))-\mu(A\ {\rm or}\ B)=\Delta_d\le 0 \\ \label{kolmogorovianfactordisjunction}
&&0 \le k_d=\mu(A)+\mu(B)-\mu(A\ {\rm or}\ B)
\end{eqnarray}
In Table 3 the values of $\Delta_d$, the `disjunction maximum rule deviation', and $k_d$, the `Kolmogorovian disjunction factor', are derived for exemplars and pairs of concepts measured by Hampton (1988a,b). For {\it Mushroom}, we have $k_d=-0.4 < 0$, which proves that a Kolmogorovian probability model is not possible.

If this type of highly non-classical overextension for disjunction only occurred for {\it Mushroom}, it would be considered an error of the experiment. However, Table 3 shows that it appears in many other exemplars as well, in only slightly less strong ways. {\it Parsley, Olive, Root Ginger, Acorn, Garlic, Almond}, all follow the same pattern as {\it Mushroom}. This means that for all of these exemplars, the answers of a substantial number of participants have invariably given rise to a kind of behavior that is highly strange from the point of view of classical logic. While the participants classified these items as `not a member of {\it Fruits}' and `not a member of {\it Vegetables}', they did classify them as `a member of {\it Fruits or Vegetables}'. It should also be noted that inequality (\ref{kolmogorovianfactordisjunction}) is violated strongly for each of these exemplars, which means that the data cannot be modeled within a classical probability structure (see Table 3). We believe that the explanation for this highly non-classical logic behavior is that the participants considered the exemplars listed above to be characteristic of the newly emerging concept {\it Fruits or Vegetables}, as a concept specially attractive for exemplars `tending to raise doubts as to whether they are fruits or vegetables'. One very clear example is {\it Tomato}, with weight 0.7 for {\it Fruits}, weight 0.7 for {\it Vegetables} and weight 1 for {\it Fruits or Vegetables}, because many indeed will doubt whether {\it Tomato} is a fruit or a vegetable. 

Of course, any convincing underpinning of our hypothesis of `the emergent concept being at the origin of over and underextension' would require more data than those of the pair of concepts {\it Fruits} and {\it Vegetables}. We therefore verified the hypothesis in great detail for all of Hampton (1988a,b) data, on disjunction and conjunction, finding that for all pairs there were several exemplars that convincingly matched our hypothesis. They are listed in Table 3. {\it Ashtray} tested for {\it Furniture}, with weight 0.7, and {\it Household Appliances}, with weight 0.3, and their disjunction {\it Furniture or Household Appliances}, with weight 0.25, is doubly underextended. We believe this to be due to {\it Ashtray} not at all being considered a member of a concept collecting items that `one can doubt about whether it is a piece of furniture or a household appliance'. A similar explanation can be given for the double underextension of the exemplars {\it Discus Throwing}, {\it Bicycle Pump}, {\it Goggles}, {\it Rat}, {\it Diving Mask} and {\it Underwater} (see Table 3).

Let us now apply our hypothesis about the 
 emergence of a new concept to the case of conjunction. We will start with {\it Mint}, tested by Hampton (1988a) for the conjunction, and yielding double overextension (see Table 3). We already considered {\it Mint} in section \ref{glimmers} and \ref{inklings}, and again our explanation is that the human mind -- following the same dynamics -- is persuaded primarily by the newly emergent concept {\it Food and Plant}. It evaluates whether {\it Mint} is a strong member of this new concept, and in doing so it does not evaluate its membership of {\it Food} or its membership of {\it Plant} in the way that {\it Food and Plant} is considered a classical logical conjunction of {\it Food} and of {\it Plant}. There is an abundance of data showing overextension for the conjunction of pairs tested in Hampton (1988a), and also a significant occurrence of double overextensions, e.g. {\it TV}, {\it Tree House} and {\it Course Liner} (see Table 3).

While the systematic presence of underextension for disjunction and of overextension for conjunction strongly support our hypothesis about the dominant dynamics of emergence of a new concept as compared to the dynamics of classical logical reasoning, it is only part of our argumentation. The other part is made up of what is more like the pieces of a puzzle containing structural and mathematical elements. In this respect, two words jump to mind, namely `average' and `Fock space'. In the following, we will briefly make a case for this second part of our argumentation.

Firstly, with regard to the word `average', let us consider the basic interference equations (\ref{membershipweightinterference}) and (\ref{muAorB}). We see that the average ${\mu(A)+\mu(B) \over 2}$ figures prominently in both. For (\ref{muAorB}), the value of $\mu(A\ {\rm or}\ B)_k$ is even given by the average plus the quantum interference term $\Re\langle A|M_k|B\rangle$, and this is so for a standard pure quantum interference equation. (\ref{membershipweightinterference}) includes an additional term $m^2\mu(A)\mu(B)$, which makes this formula not a pure interference formula. This is due to the other of the above two terms, `Fock space', and we will now explain why. If we accept our basic hypothesis that the dominant dynamics of reasoning is `emergence' and that classical logical reasoning is only secondary, it introduces `a whole new ball game'. Indeed, overextension, underextension, the guppy effect, the disjunction effect, and the conjunction fallacy -- they have all been identified assuming that classical logical reasoning is the default. As such, overextension for conjunction means `higher than the minimum', and underextension for disjunction means `lower than the maximum'. However, if we consider the dominant reasoning to be `emergence', in both cases -- conjunction as well as disjunction -- `the average' is the value that acts as a gauge. Both conjunction and disjunction fluctuate around the average, and this `fluctuation around the average' is interference. So has classical logical reasoning completely disappeared from this whole new ball game? It has not, and this is where Fock space comes in. 

Fock space is traditionally used in quantum field theory. For a quantum field, the number of entities is commonly not an actuality, i.e. not a fixed quantity. This also means that the state of a quantum field is usually a superposition of states with different fixed numbers of entities. We believe that this is exactly the situation we have identified for the combination of concepts. When two concepts combine, the human mind conceives of this combination partly as an entirely new concept -- we have used the term `emergence' to refer to this part -- and partly as an algebraic combination of two concepts, and the rules of the used algebra are quantum logic. It is in this `quantum logic' part that we find back a probabilistic version of classical logic. The human mind, however, does not choose one of these ways, but considers both together, or more correctly, a superposition of both. This superposition state can be described as a vector of Fock space. We refer to the last section of appendix A for a mathematical definition of Fock space, and to Aerts (2009a) for a detailed construction of the Fock space for two combined concepts. 

As explained in the last section of appendix A, Fock space
consists of two sectors. In the way we used it to model the combination of concepts, conjunction as well as disjunction, in sector 1, pure interference is modeled, i.e. weights fluctuate around the average, like waves. Sector 2 is a tensor product Hilbert space, and here the combination is modeled such that a probabilistic version of classical logic, more specifically quantum logic, appears as a modeling of a situation with two identical exemplars. Let us illustrate this with the following example. Consider {\it Tomato}, for {\it Fruits or Vegetables}. In sector 2 of Fock space, two identical exemplars of {\it Tomato}, hence {\it Tomato} and {\it Tomato}, are considered. One is confronted with {\it Fruits} and the other one with {\it Vegetables}. If one of these confrontations leads to acknowledgement of membership, the disjunction is satisfied. If both confrontations lead to acknowledgement of membership, the conjunction is satisfied. We can easily recognize the calculus of truth tables from classical logic in the above dynamics, except that things are probabilistic or fuzzy. 
Having sketched the dynamics in sector 2 of Fock space, we can readily see how different it is from the emergence dynamics in sector 1 of Fock space. Indeed, `quantum emergent thought', which consists in reflecting whether {\it Tomato} is a member of the new concept {\it Fruits or Vegetables}, is a completely different dynamics of thought than `quantum logical thought', i.e. the quantum probabilistic version of classical logical thought, which consists in considering two identical exemplars {\it Tomato} and {\it Tomato}, reflecting on the question whether the one is a member of {\it Fruits} and whether the other is a member of {\it Vegetables}, and then ruling for disjunction as prescribed by logic.

Also the entanglement of {\it The Animal Acts} can be analyzed now. {\it Animal} and {\it Acts} are modeled in two Hilbert spaces, and {\it The Animal Acts} is modeled in their tensor product, such that a non product vector of this tensor product can be used to describe the state of entanglement. This tensor product can be identified with a sector 2 of a Fock space. The entanglement is due to the human mind choosing exemplars of the newly emergent concept {\it The Animal Acts} in such a way that the probability weights of collapse of the state of {\it The Animal Acts} to the states of these exemplars, are not combinations of weights corresponding to exemplars that are chosen for the concepts {\it Animal} and {\it Acts} apart. This is exactly the way entanglement appears for quantum particles (Aerts \& Sozzo, 2011). 

Fock space itself is the direct sum of its two sectors. Our modeling of human reasoning is mathematically situated in the whole of Fock space, and hence in our approach human reasoning is a superposition of `emergent reasoning' and `logical reasoning' (Aerts 2009a, Aerts \& D'Hooghe, 2009). In (\ref{membershipweightinterference}), we introduced the Fock space equation for the conjunction, so that we can now explain that $m^2$ and $n^2$ are the weights with respect to sectors 2 and 1 of Fock space, i.e. the weights of `logical thought' (this is $m^2$), and `emergent thought' (this is $n^2$), in human thought as quantum superposition of both. The general Fock space equation for disjunction is
\begin{equation} \label{membershipweightinterference02}
\mu(A\ {\rm or}\ B)=m^2(\mu(A)+\mu(B)-\mu(A)\mu(B))+n^2({\mu(A)+\mu(B) \over 2}+\Re\langle A|M|B\rangle)
\end{equation}
where $m^2$ and $n^2$ play the same role, the weights of `quantum logical thought' and `quantum emergent thought' in human thought as quantum superposition of both. Interestingly, the Fock space equation for disjunction can be derived from the Fock space equation for conjunction introduced in section \ref{inklings} by applying the de Morgan laws to the `logical reasoning' sector of Fock space. Indeed, (\ref{membershipweightinterference02}) can be obtained from (\ref{membershipweightinterference}) if we replace $\mu(A)\mu(B)$ by $1-(1-\mu(A))(1-\mu(B)))=\mu(A)+\mu(B)-\mu(A)\mu(B)$, which is what we would expect also intuitively from the perspective introduced in our modeling.
This also means that the de Morgan laws are not satisfied when both `logical reasoning' and `emergent reasoning' are taken into account, which makes it possible to explain the situation of `borderline cases', as we make explicit in the following.

For the disjunction case, we have $\mu(A) \le \mu(A)+\mu(B)-\mu(A)\mu(B)$ and $\mu(B) \le \mu(A)+\mu(B)-\mu(A)\mu(B)$, from which follows that ${\mu(A)+\mu(B) \over 2} \le \mu(A)+\mu(B)-\mu(A)\mu(B)$. In the absence of interference, (\ref{membershipweightinterference02}) expresses that $\mu(A\ {\rm or}\ B)\in [{\mu(A)+\mu(B) \over 2},\mu(A)+\mu(B)-\mu(A)\mu(B)]$. Hence, values outside of this interval need a genuine interference contribution for (\ref{membershipweightinterference02}) to have a solution. We introduce
\begin{equation}
0 \le f_d=\min(\mu(A\ {\rm or}\ B)-{\mu(A)+\mu(B) \over 2},\mu(A)+\mu(B)-\mu(A)\mu(B)-\mu(A\ {\rm or}\ B))
\end{equation}
as the criterion for a possible solution without the need for genuine interference. The values of $f_d$ are in Table 3. 

What happens in sector 1 of Fock space is in many cases dominant to what happens in sector 2, and this can also be verified experimentally. Comparing the correlations of the Hampton (1988a,b) data with (i) the average, (ii) the maximum, (iii) the minimum, we find that, for the conjunction, the correlations for each of the pairs with the average are substantially higher than those with the minimum, and that, for the disjunction, the correlations for each of the pairs with the average are substantially higher than those with the maximum. In Aerts et al. (2012) we have given specifics on these correlations that illustrate the above. This concludes our argumentation for the presence in human thought as 
a superposition of a dominant dynamics of emergent thought and a secondary dynamics of logical thought.

At first sight it may seem that our quantum model for combined concepts does not incorporate order effects, which are known to exist experimentally. More concretely, experiments on the combination $A\ {\rm and}\ B$ will often lead to different data than experiments on the combination $B\ {\rm and}\ A$, and equally experiments on $A\ {\rm or}\ B$ can have different outcomes than experiments on $B\ {\rm or}\ A$. However, order effects can be modeled without problems in our approach, because in the first sector of Fock space, although $A\ {\rm and}\ B$ and $B\ {\rm and}\ A$ are described by the same superposition state, the phase of this state is different, leading to different interference angles, and hence different values for the collapse probabilities. Analogously, $A\ {\rm or}\ B$, although described by the same state as $B\ {\rm or}\ A$, generally leads to different interference angles. This is how the first sector of Fock space copes in a natural way with order effects. Intuitively we can understand why this is so. As a new emergent concept, $A\ {\rm and}\ B$ will as a rule indeed be different from $B\ {\rm and}\ A$, and equally, as a new emergent concept, $A\ {\rm or}\ B$ will be different from $B\ {\rm or}\ A$.

An interesting idea was suggested to us by one of the referees. Our modeling can be applied to the situation of `borderline cases' (see, e.g., Alxatib \& Pelletier, 2011; Blutner et al., 2012). Consider, e.g., the concept {\it A Tall Man}. For concepts of this type, the boundary between `tall' and `not tall' is not clearly defined, nor can this lack of clarity be removed by specifying the exact height of specific exemplars of the concept. As a consequence, it is well possible to find exemplars both of whose membership weights with respect to the concepts {\it A Tall Man} and {\it A Not Tall Man} are different from zero, which phenomenon is known as the {\it borderline contradiction}. The quantum modeling scheme developed in the present paper can overcome this difficulty, since it admits the possible presence of `borderline contradictions' from the very beginning. Moreover, it provides precise predictions with respect to the estimated membership weights. We plan in the next future to perform an experimental test for conceptual combinations of the `borderline' type, and to compare the predictions of our quantum interference model with a recent finding that explains `borderline contradictions' as a quantum interference phenomenon (Blutner et al., 2012).

In the following we comment on some of the questions asked by the referees because they may be of interest to other readers as well. One question was `Why does the Hilbert space we constructed in section \ref{interferencesuperposition} have 25 dimensions, while there are 24 exemplars at play?' As explained in appendix A, for our modeling it is the number of orthogonal projections of the spectral family that coincides with the number of outcomes of an experiment. In the case of the model constructed in section \ref{interferencesuperposition}, the number of outcomes of the experiment is indeed 24, each exemplar can be chosen by the participants, and this is also the number of orthogonal projection operators of the spectral family used in the model. Since we consider the disjunction of two concepts, which state is represented by the normalized superposition of two orthogonal vectors $|A\rangle$ and $|B\rangle$ representing the states of the concepts $A$ and $B$, the projection operators of the spectral family project form a space of dimension two, the subspace generated by $|A\rangle$ and $|B\rangle$. We could have constructed a solution with a Hilbert space of dimension 48, taking a two-dimensional space for each of the range spaces of the orthogonal projection operators. A solution can, however, be found whenever one two-dimensional subspace is allowed, and this is the one we presented in this article. Another question was `Why have you chosen to describe conjunction as well as disjunction in the way you do, rather than by following the quantum logical procedure, for instance, where disjunction would be linked to the join of subspaces, and conjunction to the meet of such subspaces?' This interesting question allows us to point out a conceptual aspect of our modeling that is not self-evident. It is true that in quantum logic the disjunction is connected to the join of subspaces, and the conjunction, to the meet. However, quantum logic then refers to `propositions' or, more concretely, `properties' of a quantum entity. The quantum entity itself, also in quantum logic, is described by a one-dimensional subspace, a ray, of the lattice of closed subspaces of the Hilbert space. In our approach, we consider the concept itself as an entity, and also a disjunction of two concepts, as well as a conjunction of two concepts, is again a concept, hence an entity. All of these need to be described by one-dimensional subspaces, rays, also following quantum logic. This means that our approach is compatible also with a quantum logical approach. By the way, a concept too has properties or features, and these properties will be described by closed subspaces also in our approach. Disjunctions and conjunctions, as new properties, will then be described by joins and meets of the corresponding subspaces, respectively. 

We end this section with a question that we put forward ourselves. `Why is it that, if emergent thought is dominant over logical thought, logical thought has been considered as the default all this time?'
Classical logic pretends to model `valid reasoning'. What the majority of participants in Hampton (1988b) did in considering {\it Mushroom} was emergent thought. Is emergent thought not valid reasoning? Classical logic is fruitfully applied in mathematics, without leading to situations comparable to the {\it Mushroom} case. Could this hint at an answer to the above questions? We think so. Classical logic has been discriminative towards certain types of propositions from the start. The principle of the excluded middle, one of the three basic axioms of classical logic, states  that `for any proposition either it is true or its negation is'. This principle cuts away the potential for emergence, and it is well-known that it is not valid for propositions concerning quantum entities. Already the Ancient Greeks put forward a very simple proposition that did not satisfy this principle, the liar paradox proposition, i.e. the proposition `this proposition is false'. In this sense, it is not a coincidence that we described the liar paradox as a cognitive situation able to be modeled by quantum theory (Aerts et al., 1999). The proposition `this proposition is true', although much less vicious, does not satisfy the principle of the excluded middle either. If it is true, it is true, and if it is false, it is false. Hence, the `being true' and the `being false' depend directly on the hypothesis made. This is contextuality in its purest form, and we have analyzed its consequences. 

With respect to the principle of the excluded middle, there have been serious doubts about its validity throughout history. Aristotle pondered about propositions that referred to future events, such as `There will be a sea battle tomorrow' (Aristotle, -350). The famous example of the sea battle brings us closer to understanding the question about the predominance of logical thought. We all know that `the future is about potentialities and not actualities'. However, we believe that `the present is about actualities and not potentialities'. This is too simple a view of the present. Once context plays an essential role with respect to the values of interest, these values start to refer to potentialities, and no longer to actualities. To respond to the statement, `being a smoker or not', we can still consider only actualities, but this no longer holds for statements such as `being in favor or against the use of nuclear energy', because for this type of statement context plays an essential role (Aerts \& Aerts, 1995). This is the reason why quantum structure, which is mathematically capable of coping with potentialities, comes into play whenever the human mind needs to assess values. This is the reason why also emergent thought is valid reasoning.

\vspace{-0.4cm}
\section{Conclusion}
\vspace{-0.4cm}

We presented our approach to concepts modeling and the way we elaborated a description of concept combinations by using the mathematical formalism of quantum theory 

We explained how several findings in concept research, including the `graded nature of exemplars', the `guppy effect' and the `overextension' and `underextension' of membership weights, led us to recognize the need for quantum modeling and, more specifically, for an approach where `concepts can be in different states' and `change states under the influence of context'. We positioned our approach with respect to the main traditional concept theories, prototype theory, exemplar theory, and theory theory (section \ref{glimmers}).

We illustrated our quantum modeling approach by providing a description of the overextension for conjunctions of concepts measured by Hampton (1988a) as an effect of quantum interference. We analyzed in depth what makes quantum interference modeling superior to classical probabilistic or fuzzy set modeling, and pointed out the essential role of complex numbers (section \ref{inklings}).

In particular, we considered the concept combination {\it The Animal Acts}, and explained how we performed an experiment based on this concept combination to test Bell's inequalities, and how this experimented resulted in a violation of the inequalities. In this way we proved the presence of quantum entanglement in concept combination (section \ref{entanglement}). 

Furthermore, we considered interference and superposition in the disjunction {\it Fruits or Vegetables}, at the same time showing that quantum interference patterns naturally appear whenever suitable exemplars of this disjunction are taken into account. We have calculated the graphics revealing the interference pattern of {\it Fruits or Vegetables} and compared them with the interference of light in a double slit quantum experiment (section \ref{interferencesuperposition}).

Finally, we enlightened, again by means of examples, that also emergence occurs in conceptual processes, and we gathered arguments of an experimental and theoretical nature to put forward our main explanatory hypothesis, namely that human thought is the quantum superposition of `quantum emergent thought' and `quantum logical 
thought', and that our quantum modeling approach applied in Fock space enables this general human thought to be modeled (section \ref{emergence}).

\bigskip
\noindent
\large
{\bf Appendices}
\normalsize

\appendix
\vspace{-0.4cm}
\section{Quantum Theory for Modeling}
\vspace{-0.4cm}
When quantum theory is applied for modeling purposes, each entity considered -- in our case a concept -- has a corresponding complex Hilbert space ${\cal H}$, which is a vector space over the field ${\mathbb C}$ of complex numbers, equipped with an inner product $\langle \cdot |  \cdot \rangle$, that maps two vectors $\langle A|$ and $|B\rangle$ to a complex number $\langle A|B\rangle$. We denote vectors using the bra-ket notation introduced by Paul Adrien Dirac, one of the founding fathers of quantum mechanics (Dirac, 1958). Vectors can be kets, denoted $\left| A \right\rangle $, $\left| B \right\rangle$, or bras, denoted $\left\langle A \right|$, $\left\langle B \right|$. The inner product between the ket vectors $|A\rangle$ and $|B\rangle$, or the bra-vectors $\langle A|$ and $\langle B|$, is realized by juxtaposing the bra vector $\langle A|$ and the ket vector $|B\rangle$, and $\langle A|B\rangle$ is also called a bra-ket, and it satisfies the following properties: (i) $\langle A |  A \rangle \ge 0$; (ii) $\langle A |  B \rangle=\langle B |  A \rangle^{*}$, where $\langle B |  A \rangle^{*}$ is the complex conjugate of $\langle A |  B \rangle$; $\langle A |(z|B\rangle+t|C\rangle)=z\langle A |  B \rangle+t \langle A |  C \rangle $, for $z, t \in {\mathbb C}$,
where the sum vector $z|B\rangle+t|C\rangle$ is called a `superposition' of vectors $B\rangle$ and $C\rangle$ in the quantum jargon. From (ii) and (iii) follows that it is linear in the ket and anti-linear in the bra, i.e. $(z\langle A|+t\langle B|)|C\rangle=z^{*}\langle A | C\rangle+t^{*}\langle B|C \rangle$. We recall that the absolute value of a complex number is defined as the square root of the product of this complex number times its complex conjugate. Hence we have $|\langle A| B\rangle|=\sqrt{\langle A|B\rangle\langle B|A\rangle}$. We define the `length' of a ket (bra) vector $|A\rangle$ ($\langle A|$) as $|| |A\rangle ||=||\langle A |||=\sqrt{\langle A |A\rangle}$. A vector of unitary length is called a `unit vector'. We say that the ket vectors $|A\rangle$ and $|B\rangle$ are `orthogonal' and write $|A\rangle \perp |B\rangle$ if $\langle A|B\rangle=0$. We have introduced the necessary mathematics to describe the first modeling rule of quantum theory, which is the following. 

\medskip
\noindent{\it First quantum modeling rule:} A state of an entity -- in our case a concept -- modeled by quantum theory is represented by a ket vector $|A\rangle$ with length 1, i.e. $\langle A|A\rangle=1$.

\medskip
\noindent
An orthogonal projection $M$ is a linear function on the Hilbert space, hence $M: {\cal H} \rightarrow {\cal H}, |A\rangle \mapsto M|A\rangle$, which is Hermitian and idempotent, which means that for $|A\rangle, |B\rangle \in {\cal H}$ and $z, t \in {\mathbb C}$ we have (i) $M(z|A\rangle+t|B\rangle)=zM|A\rangle+tM|B\rangle$ (linearity); (ii) $\langle A|M|B\rangle=\langle B|M|A\rangle$ (hermiticity); and (iii) $M \cdot M=M$ (idempotency). The identity, mapping each vector on itself, is a trivial orthogonal projection, denoted $\mathbbmss{1}$. We say that two orthogonal projections $M_k$ and $M_l$ are orthogonal, if each vector contained in $M_k({\cal H})$ is orthogonal to each vector contained in $M_l({\cal H})$, and we write in this case $M_k \perp M_l$. A set of orthogonal projection operators $\{M_k\ \vert k=1,\ldots,n\}$ is called a spectral family, if all projectors are mutually orthogonal, i.e. $M_k \perp M_l$ for $k \not= l$, and their sum is the identity, i.e. $\sum_{k=1}^nM_k=\mathbbmss{1}$. This gives us the necessary mathematics to describe the second modeling rule.

\medskip
\noindent
{\it Second quantum modeling rule:} A measurable quantity of an entity -- in our case a concept -- modeled by quantum theory, and having a set of possible real values $\{a_1, \ldots, a_n\}$ is represented by a spectral family $\{M_k\ \vert k=1, \ldots, n\}$ in the following way. If the entity is in a state represented by the vector $|A\rangle$, this state is changed to a state represented by one of the vectors $M_k|A\rangle/\|M_k|A\rangle\|$, with probability $\langle A|M_k|A\rangle$. In this case the value of the quantity is $a_k$,and the change of state taking place is called collapse in the quantum jargon.

\medskip
\noindent
The tensor product ${\cal H}_{A} \otimes {\cal H}_{B}$ of two Hilbert spaces ${\cal H}_{A}$ and ${\cal H}_{B}$ is the Hilbert space generated by the set $\{|A\rangle_i \otimes |B\rangle_j\}$, where $|A\rangle_i$ and $|B\rangle_j$ are vectors of ${\cal H}_{A}$ and ${\cal H}_{B}$, respectively, which means that a general vector of this tensor product is of the form $\sum_{ij}|A\rangle_i \otimes |B\rangle_j$. This gives us the necessary mathematics to introduce the third modeling rule.

\medskip
\noindent
{\it Third quantum modeling rule:} A state $p$ of a compound entity -- a combined concept -- is represented by a unit vector $|C\rangle$ of the tensor product ${\cal H}_{A} \otimes {\cal H}_{B}$ of the two Hilbert spaces ${\cal H}_{A}$ and ${\cal H}_{B}$ containing the vectors that represent the states of the component entities -- concepts.

\medskip
\noindent
The above means that we have $|C\rangle=\sum_{ij}|A\rangle_i \otimes |B\rangle_j$, where $|A\rangle_i$ and $|B\rangle_j$ are unit vectors of ${\cal H}_{A}$ and ${\cal H}_{B}$, respectively. We say that the state $p$ represented by $|C\rangle$ is a product state if it is of the form $|A\rangle \otimes |B\rangle$ for some $|A\rangle \in {\cal H}_{A}$ and $|B\rangle \in {\cal H}_{B}$. Otherwise, $p$ is called an `entangled state'.

\medskip
\noindent
Fock space is a specific type of Hilbert space, originally introduced in quantum field theory. For most states of a quantum field the number of identical quantum entities is not an actuality, i.e. predictable quantity. Fock space copes with this situation in allowing its vectors to be superpositions of vectors pertaining to sectors for fixed numbers of identical quantum entities. Such a sector, describing a fixed number of $j$ identical quantum entities, is of the form ${\cal H}\otimes \ldots \otimes{\cal H}$ of the tensor product of $j$ identical Hilbert spaces ${\cal H}$. Fock space $F$ itself is the direct sum of all these sectors, hence
\begin{equation} \label{fockspace}
F=\oplus_{k=1}^j\otimes_{l=1}^k{\cal H}
\end{equation}
For our modeling we have only used Fock space for the `two' and `one quantum entity' case, hence $F={\cal H}\oplus({\cal H}\otimes{\cal H})$. This is due to considering only combinations of two concepts. A unit vector $|F\rangle \in F$ is then written as $|F\rangle = ne^{i\gamma}|C\rangle+me^{i\delta}(|A\rangle\otimes|B\rangle)$, where $|A\rangle, |B\rangle$ and $|C\rangle$ are unit vectors of ${\cal H}$, and such that $n^2+m^2=1$. For combinations of $j$ concepts, the general form of Fock space expressed in equation (\ref{fockspace}) will have to be used.

\vspace{-0.4cm}
\small
\section*{References}
\begin{description}
\vspace{-0.2cm}
\setlength{\itemsep}{-2mm}
\item Aerts, D. (1982a). Description of many physical entities without the paradoxes encountered in quantum mechanics. {\it Foundations of Physics 12}, 1131--1170.

\item Aerts, D. (1982b). Example of a macroscopical situation that violates Bell inequalities. {\it Lettere al Nuovo Cimento 34}, 107--111.

\item Aerts, D. (1986). A possible explanation for the probabilities of quantum mechanics. {\it Journal of Mathematical Physics 27}, 202--210.

\item Aerts, D. (1991). A mechanistic classical laboratory situation violating the Bell inequalities with 2$\sqrt{2}$, exactly 'in the same way' as its violations by the EPR experiments. {\it Helvetica Physica Acta 64}, 1--23.

\item Aerts, D. (1999). Foundations of quantum physics: a general realistic and operational approach. {\it International Journal of Theoretical Physics 38}, 289--358.  

\item Aerts, D. (2009a). Quantum structure in cognition. {\it Journal of Mathematical Psychology 53}, 314--348.

\item Aerts, D. (2009b). Quantum particles as conceptual entities: A possible explanatory framework for quantum theory. {\it Foundations of Science 14}, 361--411. 

\item Aerts, D., \& Aerts, S. (1995). Applications of quantum statistics in psychological studies of decision processes. {\it Foundations of Science 1}, 85-97.

\item Aerts, D., Aerts, S., Broekaert, J., \& Gabora, L. (2000). The violation of Bell inequalities in the macroworld. {\it Foundations of Physics 30}, 1387--1414.

\item Aerts, D., Aerts, S., Coecke, B., D'Hooghe, B., Durt, T. and Valckenborgh, F. (1997). A model with varying fluctuations in the measurement context. In M. Ferrero and A. van der Merwe (Eds.), {\it New Developments on Fundamental Problems in Quantum Physics} (pp. 7-9). Dordrecht: Springer.

\item Aerts, D., Aerts, S., \& Gabora, L. (2009). Experimental evidence for quantum structure in cognition. {\it Quantum Interaction. Lecture Notes in Computer Science 5494}, 59--70.

\item Aerts, D., Broekaert, J. and Smets, S. (1999). A quantum structure description of the liar paradox. {\it International Journal of Theoretical Physics 38}, 3231--3239.

\item Aerts, D., Broekaert, J., Gabora, L. and Veloz, T. (2012). The Guppy Effect as Interference. In the {\it Proceedings of the sixth International Symposium on Quantum Interaction}, 27-29 June 2012, Paris, France.

\item Aerts, D., Coecke, B., \& Smets, S. (1999). On the origin of probabilities in quantum mechanics: creative and contextual aspects. In Cornelis, G., Smets, S., \& Van Bendegem, J. P. (Eds.), {\it Metadebates on Science} (pp. 291-302). Dordrecht: Springer.

\item Aerts, D. and Czachor, M. (2004). Quantum aspects of semantic analysis and symbolic artificial intelligence. {\it Journal of Physics A: Mathematical and Theoretical 37}, L123--L132. 

\item Aerts, D., \& D'Hooghe, B. (2009). Classical logical versus quantum conceptual thought: Examples in economy, decision theory and concept theory. {\it Quantum Interaction. Lecture Notes in Artificial Intelligence 5494}, 128--142.

\item Aerts, D. and Durt, T. (1994). Quantum, classical and intermediate, an illustrative example. {\it Foundations of Physics 24}, 1353--1369.

\item Aerts, D., Durt, T., Grib, A., Van Bogaert, B. and Zapatrin, A. (1993). Quantum structures in macroscopical reality. {\it International Journal of Theoretical Physics 32}, 489--498. 

\item Aerts, D., \& Gabora, L. (2005a). A theory of concepts and their combinations I: The structure of the sets of contexts and properties. {\it Kybernetes 34}, 167--191.

\item Aerts, D., \& Gabora, L. (2005b). A theory of concepts and their combinations II: A Hilbert space representation. {\it Kybernetes 34}, 192--221.

\item Aerts, D., \& Sozzo, S. (2011). Quantum structure in cognition. Why and how concepts are entangled. {\it Quantum Interaction. Lecture Notes in Computer Science 7052}, 116--127.

\item Aerts, D., \& Van Bogaert, B. (1992). Mechanistic classical laboratory situation with a quantum logic structure. {\it International Journal of Theoretical Physics 31}, 1839--1848.

\item Alxatib, S., \& Pelletier, J. (2011). On the psychology of truth gaps. In Nouwen, R., van Rooij, R., Sauerland, U., \& Schmitz, H.-C. (Eds.), {\it Vagueness in Communication} (pp. 13--36). Berlin, Heidelberg: Springer-Verlag.

\item Aristoteles (-350). On Interpretation.

\item Aspect, A., Grangier, P., \& Roger, G. (1982). Experimental realization of Einstein-Podolsky-Rosen-Bohm gedankenexperiment: A new violation of Bell's inequalities. {\it Physical Review Letters 49}, 91--94.

\item Bell, J. S. (1964). On the Einstein-Podolsky-Rosen paradox. {\it Physics 1}, 195--200.

\item Blutner, R., Pothos, E. M., \& Bruza, P. (2012). A quantum probability perspective on borderline vagueness. {\it Topics in Cognition Science} (in print).

\item Bruza, P. D., Kitto, K., McEvoy, D., McEvoy, C. (2008). Entangling words and meaning. In {\it Proceedings of the Second Quantum Interaction Symposium}. Oxford: Oxford University, 118--124.

\item Bruza, P. D., Kitto, K., Nelson, D., McEvoy, C. (2009). Extracting spooky-activation-at-a-distance from considerations of entanglement. {\it Lecture Notes in Computer Science 5494}, 71--83.

\item Busemeyer, J. R., Wang, Z. and Townsend, J. T. (2006). Quantum dynamics of human decision-making. {\it Journal of Mathematical Psychology 50}, 220--241.

\item Busemeyer, J. R., Pothos, E. M., Franco, R. \& Trueblood, J. S. (2011). A quantum theoretical explanation for probability judgment errors. {\it Psychological Review 118}, 193--218. 

\item Cardano, G. (1545). Artis Magnae, Sive de Regulis Algebraicis (also known as Ars magna). Nuremberg. 

\item Clauser, J. F., Horne, M. A., Shimony, A., \& Holt, R. A. (1969). Proposed experiment to test local hidden-variable theories. {\it Physical review Letters 23}, 880--884.

\item Deerwester, S., Dumais, S. T., Furnas, G. W., Landauer,   T. K. \& Harshman,  R. (1990). Indexing by Latent Semantic Analysis. {\it Journal of the American Society for Information Science 41} 391--407.

\item Dirac, P. A. M. (1958). {\it Quantum mechanics}, 4th ed. London: Oxford University Press.

\item Franco, R. (2009). The conjunctive fallacy and interference effects. {\it Journal of Mathematical Psychology 53}, 415--422.

\item Feynman, R. (1967). {\it The Character of Physical Law}. M.I.T. Press.

\item Gabora, L., (2001). {\it Cognitive Mechanism Underlying the Origin and Evolution of Culture.} Doctoral Dissertation, Center Leo Apostel, University of Brussels.

\item Gabora, L., \& Aerts, D. (2002). Contextualizing concepts using a mathematical generalization of the quantum formalism. {\it Journal of Experimental and Theoretical Artificial Intelligence 14}, 327--358.

\item Galea, D., Bruza, P. D., Kitto, K., Nelson, D., McEvoy, C. (2011). Modelling the activation of words in human memory: The spreading activation, Spooky-Activation-at-a-Distance and the entanglement models compared. {\it Lecture Notes in Computer Science 7052}, 149--160.

\item Hampton, J. A. (1988a). Overextension of conjunctive concepts: Evidence for a unitary model for concept typicality and class inclusion. {\it Journal of Experimental Psychology: Learning, Memory, and Cognition 14}, 12--32.

\item Hampton, J. A. (1988b). Disjunction of natural concepts. {\it Memory \& Cognition 16}, 579--591.

\item Hampton, J. A. (1997). Conceptual combination: Conjunction and negation of natural concepts. {\it Memory \& Cognition 25}, 888--909.

\item Kamp, H., \& Partee, B. (1995). Prototype theory and compositionality. {\it Cognition 57}, 129--191. 

\item Jauch, J. M. (1968). {\it Foundations of Quantum Mechanics}. Addison-Wesley Publishing Company: Reading, Mass.

\item Kolmogorov, A. N. (1933). {\it Grundbegriffe der Wahrscheinlichkeitrechnung}, Ergebnisse Der Mathematik; translated as {\it Foundations of Probability}. New York: Chelsea Publishing Company, 1950.

\item Lambert Mogiliansky, A., Zamir, S., \&  Zwirn, H. (2009). Type indeterminacy: A model of the KT(Kahneman-Tversky)-man. {\it Journal of Mathematical Psychology 53}, 349--36.

\item Laplace, P. S. (1820). Th\'eorie analytique des probabilit\'es. Paris : Mme Ve Courcier.

\item Li, Y., \& Cunningham, H. (2008). Geometric and quantum methods for information retrieval. {\it ACM SIGIR Forum 42}, 22--32. 

\item Mackey, G. (1963). {\it Mathematical Foundations of Quantum Mechanics}, Reading: W. A. Benjamin.

\item Melucci, M. (2008). A basis for information retrieval in context. {ACM Transactions of Information Systems 26}, 1--41.

\item Murphy, G. L., \& Medin, D. L. (1985). The role of theories in conceptual coherence. {\it Psychological Review 92}, 289-–316.

\item  Nelson, D. L., McEvoy, C. L. (2007). Entangled associative structures and context. In Bruza, P. D., Lawless, W., van Rijsbergen, C. J., Sofge, D. (Eds.) {\it Proceedings of the AAAI Spring Symposium on Quantum Interaction}. Menlo Park: AAAI Press.

\item Nosofsky, R. (1988). Exemplar-based accounts of relations between classification, recognition, and typicality. {\it Journal of Experimental Psychology: Learning, Memory, and Cognition 14}, 700–-708.

\item Nosofsky, R. (1992). Exemplars, prototypes, and similarity rules. In Healy, A., Kosslyn, S., \& Shiffrin, R. (Eds.), {\it From learning theory to connectionist theory: Essays in honor of William K. Estes}. Hillsdale NJ: Erlbaum.

\item Osherson, D., \& Smith, E. (1981). On the adequacy of prototype theory as a theory of concepts. {\it Cognition 9}, 35--58.

\item Osherson, D. N., \& Smith, E. (1997). On typicality and vagueness. {\it Cognition 64}, 189--206.

\item Piron, C. (1976). {\it Foundations of Quantum Physics}, Reading Mass.: W. A. Benjamin.

\item Piwowarski, B., Frommholz, I., Lalmas, M., \& van Rijsbergen, K. (2010). What can Quantum Theory bring to IR. In Huang, J., Koudas, N., Jones, G., Wu, X., Collins-Thompson, K., \& An, A. (Eds.), {\it CIKM'10: Proceedings of the nineteenth ACM conference on Conference on information and knowledge management}.

\item Pothos, E. M. \& Busemeyer, J. R. (2009). A quantum probability explanation for violations of `rational' decision theory. {\it Proceedings of the Royal Society B 276}, 2171--2178.

\item Rosch, E. (1973). Natural categories, {\it Cognitive Psychology 4}, 328--350.

\item Rosch, E. (1978). Principles of categorization. In Rosch, E. \& Lloyd, B. (Eds.), {\it Cognition and categorization}. Hillsdale, NJ: Lawrence Erlbaum, pp. 133--179.

\item Rosch, E. (1983). Prototype classification and logical classification: The two systems. In Scholnick, E. K. (Ed.), {\it New trends in conceptual representation: Challenges to Piaget's theory?}. Hillsdale, NJ: Lawrence Erlbaum, pp. 133--159.

\item Rumelhart, D. E., \& Norman, D. A. (1988). Representation in memory. In Atkinson, R. C., Hernsein, R. J., Lindzey, G., \& Duncan, R. L. (Eds.), {\it Stevens' handbook of experimental psychology}. Hoboken, New Jersey: John Wiley \& Sons.

\item Tversky, A. \& Kahneman, D. (1983). Extension versus intuitive reasoning: The conjunction fallacy in probability judgment. {\it Psychological
Review 90}, 293--315.

\item Tversky, A., \& Shafir, E. (1992). The disjunction effect in choice under uncertainty. {\it Psychological Science 3}, 305--309. 

\item Van Rijsbergen, K. (2004) {\it The Geometry of Information Retrieval}, Cambridge, UK: Cambridge University Press.

\item  V. J. Weeden, D. L. Rosene, R. Wang, G. Dai, F. Mortazavi, P. Hagmann, J. H. Kaas, and W. I. Tseng (2012). The geometric structure of the brain fiber pathways. {\it Science 335}, 1628--1634.

\item Wang, Z., Busemeyer, J. R., Atmanspacher, H., \& Pothos, E. (this issue). The potential of quantum probability for modeling cognitive processes. {\it Topics in Cognitive Science}.

\item Widdows, D., \& Peters, S. (2003) Word vectors and quantum logic: Experiments with negation and disjunction, in {\it Mathematics of Language 8}, Indiana, IN: Bloomington, pp. 141--154.

\item Widdows, D. (2006) {\it Geometry and Meaning}, CSLI Publications, IL: University of Chicago Press.

\item Wittgenstein, L. (1953/2001). {\it Philosophical Investigations,  \S 65-71.} Blackwell Publishing.

\item Zuccon, G., \& Azzopardi, L. (2010). Using the quantum probability ranking principle to rank Interdependent documents. In Gurrin, G., He, Y., Kazai, G., Kruschwitz, U., Little, S., Roelleke, T., R{\"u}ger, S., et al. (Eds.), {\it Advances in Information Retrieval 5993}. Berlin: Springer, pp. 357--369.

\end{description}

\end{document}